%% file: main.tex
\documentclass[runningheads]{llncs}
\usepackage[T1]{fontenc}
\usepackage{graphicx}
\usepackage[breaklinks,colorlinks]{hyperref}
\usepackage{xcolor}

\hypersetup{hidelinks,
backref=true,
pagebackref=true,
hyperindex=true,
breaklinks=true,
colorlinks=true,%linkcolor=black,
urlcolor=blue,
bookmarks=true,
bookmarksopen=false,
pdftitle={Title},
pdfauthor={Author}}

\usepackage{booktabs}
\usepackage{listings}
\usepackage[section]{placeins}
\usepackage{tabularx}
\usepackage{multirow, multicol}
\usepackage{amsmath}  % For \operatorname
\usepackage{amssymb}  % For \mathcal and \mathbb
\usepackage{booktabs}
\usepackage{colortbl}    % For row colors
\usepackage{breakcites}

\input{macros}

\begin{document}

\title{
    \texorpdfstring{
        \includegraphics[width=0.031\textwidth]{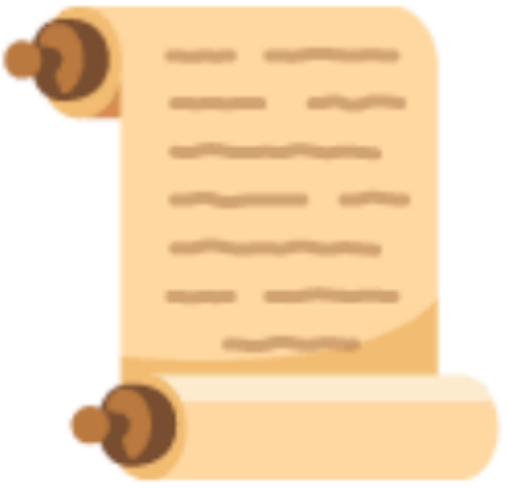}
         POEM: Precise Object-level Editing \\ via MLLM control
    }{POEM: Precise Object-level Editing via MLLM control}
}

% \title{
%     %\includegraphics[width=0.04\textwidth]{_legacy/img/VIOLET.png} % Adjust the width as needed
%     %VIOLET: VLM-guided Object-Level Image Editing and Transformation
%     \emoji{scroll} POEM: Precise Object-level Editing \\ via MLLM control\vspace{-4pt}%due to emoji
% }
% \titlerunning{VLM controller for precise mask generation in image editing}
% \titlerunning{VLM-guided Object-Level Image Editing and Transformation}
\titlerunning{Precise Object-level Editing via MLLM control}

% \author{Anonymous SCIA 2025 Submission}
% \institute{Paper ID 16}

% After submission
% \author{{Marco Schouten\inst{1,3}\orcidID{0009-0003-8243-4354}} \and
% {Mehmet Onurcan Kaya\inst{1,3}\orcidID{0009-0006-2606-3992}}
%  \and
% {Serge Belongie\inst{2,3}\orcidID{0000-0002-0388-5217
% }} \and {Dim P. Papadopoulos\inst{1,3}\orcidID{0000-0002-5278-2273}}}
% \authorrunning{M. Schouten et al.}
% \institute{Technical University of Denmark, 2800 Kongens Lyngby, Denmark
% \email{\{marscho,monka,dimp\}@dtu.dk} \and University of Copenhagen, 1172 Copenhagen, Denmark\\ \email{s.belongie@di.ku.dk} \and Pioneer Center for AI, 1350 Copenhagen, Denmark \\
% \href{https://poem.compute.dtu.dk}{https://poem.compute.dtu.dk}} 

\author{
Marco Schouten$^{1,3}$ \quad Mehmet Onurcan Kaya$^{1,3}$ \\ Serge Belongie$^{2,3}$ \quad Dim~P.~Papadopoulos$^{1,3}$\\[0.5em]
$^{1}$Technical University of Denmark \quad $^{2}$University of Copenhagen\\
$^{3}$Pioneer Centre for AI\\[0.5em]
\texttt{\small marscho@dtu.dk, monka@dtu.dk, s.belongie@di.ku.dk, dimp@dtu.dk}\\
\url{https://poem.compute.dtu.dk}
}
\authorrunning{M. Schouten et al.}


\maketitle
\input{0_abstract}

\input{1_introduction}

\input{2_related_work}
\input{3_method}

\input{4_experiments}
\input{5_conclusion}
% \input{6_appendix}

% \begin{credits}
% \subsubsection{\ackname} 

% % \subsubsection{\discintname}
% % The authors have no competing interests to declare that are relevant to the content of this article.
% \end{credits}

% \small{
% \bibliography{main.bib}

% }

\bibliographystyle{splncs04}

\input{main.bbl}
\end{document}

%% file: macros.tex
\iftrue     
\newcommand{\dimpp}[1]{\textcolor{red}{[DP: #1]}}
\newcommand{\marco}[1]{\textcolor{blue}{[MS: #1]}}
\newcommand{\onur}[1]{\textcolor{magenta}{#1}}
\else
\newcommand{\dimpp}[1]{\textcolor{red}{\noindent}}
\newcommand{\marco}[1]{\textcolor{blue}{\noindent}}
\newcommand{\onur}[1]{\textcolor{orange}{\noindent}}
\fi

\newcommand{\mypar}[1]{\vspace{0mm}\noindent\textbf{#1}}

\definecolor{sec31_yellow}{RGB}{214,182,86}
\definecolor{sec32_violet}{RGB}{150,115,166}
\definecolor{sec33_green}{RGB}{130,179,102}
\definecolor{sec35_blue}{RGB}{108,142,191}

%% file: 0_abstract.tex
\begin{abstract}
Diffusion models have significantly improved text-to-image generation, producing high-quality, realistic images from textual descriptions. Beyond generation, object-level image editing remains a challenging problem, requiring precise modifications while preserving visual coherence. Existing text-based instructional editing methods struggle with localized shape and layout transformations, often introducing unintended global changes. Image interaction-based approaches offer better accuracy but require manual human effort to provide precise guidance.  To reduce this manual effort while maintaining a high image editing accuracy, in this paper, we propose POEM, a framework for Precise Object-level Editing using Multimodal Large Language Models (MLLMs). POEM leverages MLLMs to analyze instructional prompts and generate precise object masks before and after transformation, enabling fine-grained control without extensive user input. This structured reasoning stage guides the diffusion-based editing process, ensuring accurate object localization and transformation. To evaluate our approach, we introduce VOCEdits, a benchmark dataset based on PASCAL VOC 2012, augmented with instructional edit prompts, ground-truth transformations, and precise object masks. Experimental results show that POEM outperforms existing text-based image editing approaches in precision and reliability while reducing manual effort compared to interaction-based methods.
\end{abstract}

%% file: 1_introduction.tex
\section{Introduction}

% (A)
% T2I generation, diffusion models
% Image editing. enable users to refine input images. What's the main challenge?

Recent advances in computer vision have been driven by diffusion models~\cite{rameshHierarchicalTextConditionalImage2022,rombachHighResolutionImageSynthesis2022}, which have substantially improved high-resolution text-to-image generation, producing highly realistic and diverse images from textual descriptions.
Beyond generation, image editing~\cite{lugmayrRePaintInpaintingUsing2022,mengSDEditGuidedImage2022} has emerged as a crucial application, enabling users to modify input images according to their needs while preserving realism. A challenging aspect of image editing is precise object-level modifications, such as transforming individual target objects while maintaining structural coherence. While existing techniques allow for global adjustments~\cite{brooksInstructPix2PixLearningFollow2023}, achieving fine-grained, localized edits with high accuracy remains an open research problem~\cite{kawarImagicTextBasedReal2023}.

\begin{figure}[t]
    \centering
    \includegraphics[width=\linewidth]{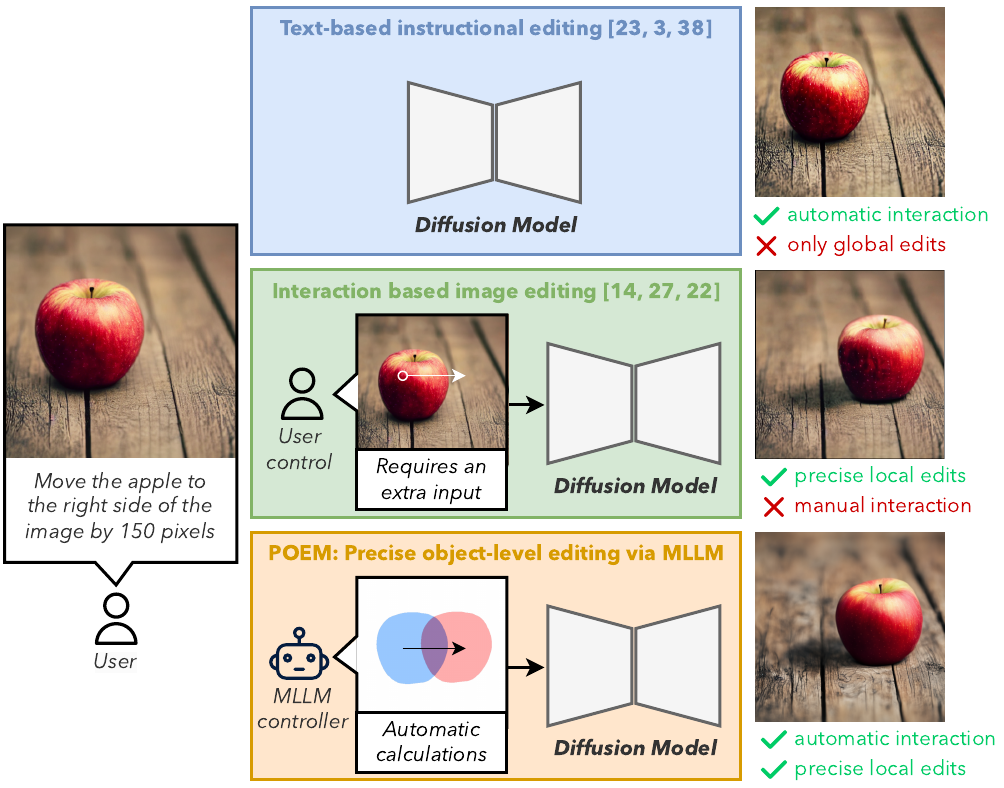}
    \caption{ \textbf{POEM.} Existing text-based instruction editing methods (top) struggle with precise object-level shape and layout edits. 
    Image interaction-based approaches (middle) perform better but require significant manual user effort. Instead, we propose (bottom) leveraging MLLMs to interpret instructional prompts and automatically generate precise object masks and numerical transformations to support image editing pipelines.} 
    \label{fig:teaser}
\end{figure}
% \FloatBarrier

%(B) 
% Existing editing. 2 categories. Text-based Instructional editing like IP2P and image interaction based editing. 
% I2P2: modifies image with single edit prompt. User-friendly. However, they struggle. why? purely rely on cross-attention text conditioning embeddings  (fig 1 top)
% 2nd strategy.  (fig 2 middle). Users are asked to provide addtional guidance in the form of image interaction (clicks, scribbles, masks, draggin points). accruate, time consuming. better but significant manual effort.

Broadly, image editing methods fall into two categories: text-based instructional editing~\cite{brooksInstructPix2PixLearningFollow2023,mengSDEditGuidedImage2022,mokadyNulltextInversionEditing2023,ruizDreamBoothFineTuning2023} and image interaction-based editing~\cite{cuiStableDragStableDragging2024,epsteinDiffusionSelfGuidanceControllable2023,gengMotionGuidanceDiffusionBased2024,lugmayrRePaintInpaintingUsing2022,mouDiffEditorBoostingAccuracy2024,nicholGLIDEPhotorealisticImage2022,parkShapeGuidedDiffusionInsideOutside2024,voynovSketchGuidedTexttoImageDiffusion2022,xieSmartBrushTextShape2023}. The former category, exemplified by InstructPix2Pix~\cite{brooksInstructPix2PixLearningFollow2023}, modifies input images based on a single edit prompt, making it efficient and user-friendly. Even though these methods have shown compelling results with global edits, they struggle with precise object-level shape transformations, often producing unintended global changes (Fig.~\ref{fig:teaser}, top). This is mainly because they purely rely on cross-attention text conditioning of a stable diffusion model~\cite{brooksInstructPix2PixLearningFollow2023,mengSDEditGuidedImage2022}.
In contrast, interaction-based approaches require users to provide additional guidance through
% image  scribbles-sketch~\cite{voynovSketchGuidedTexttoImageDiffusion2022},
precise object masks~\cite{lugmayrRePaintInpaintingUsing2022,nicholGLIDEPhotorealisticImage2022,parkShapeGuidedDiffusionInsideOutside2024,xieSmartBrushTextShape2023}, specific object modification shapes~\cite{epsteinDiffusionSelfGuidanceControllable2023} or click and drag~\cite{cuiStableDragStableDragging2024,gengMotionGuidanceDiffusionBased2024,mouDiffEditorBoostingAccuracy2024} (Fig.~\ref{fig:teaser}, middle). While these methods can localize edits accurately and improve object-level editing, they demand significant manual effort, making them less scalable.

% (C) POEM. main motivation of our method.

To address these limitations, we introduce \textbf{POEM} (\textbf{P}recise \textbf{O}bject-level \textbf{E}diting via \textbf{M}LLM control)(Fig.~\ref{fig:teaser}, bottom), 
a novel framework that decouples visual reasoning from the editor to achieve fine-grained object transformations.
% a novel framework that decouples visual reasoning from image editing to achieve fine-grained object transformations.
Instead of requiring users to provide precise image interactions, POEM leverages Multimodal Large Language Models (MLLMs) to interpret instructional prompts, generate precise object masks before and after transformation, and provide detailed image content descriptions. Inspired by recent advancements in large language models (LLMs) for complex reasoning~\cite{ganiLLMBlueprintEnabling2024,wuSelfcorrectingLLMcontrolledDiffusion2023} and MLLMs~\cite{fengLayoutGPTCompositionalVisual2023,peiSOWingInformationCultivating2024,yuPromptFixYouPrompt2024} for guiding diffusion processes, POEM ensures object localization and transformation without requiring extensive manual annotation.

Given an input image and a user edit instruction, 
POEM operates in two stages (Fig.~\ref{fig:2}). In the reasoning stage, MLLMs generate structured editing instructions, including precise segmentation masks that define object boundaries before and after the transformation. These masks then guide the editing stage, where we apply controlled modifications in the latent space of a pre-trained diffusion model. By constraining the generation process with explicitly defined regions, POEM ensures fine-grained control over object transformations, surpassing previous text-based approaches in precision and reliability.

% (E) Dataset and results
Existing datasets for image editing~\cite{yuAnyEditMasteringUnified2024,zhangMagicBrushManuallyAnnotated2024} evaluate generic editing instructions, but they fail to capture the nuanced variations and fine details that are critical when assessing object shape edits.
To address this gap and validate our method, we introduce a novel dataset, VOCEdits, by augmenting the training set of PASCAL VOC 2012~\cite{everingham2015pascal}  with instructional edits and precise ground-truth object masks for before-and-after transformations. Our dataset enables a more rigorous evaluation of our framework's ability to handle specific edit requests, which existing datasets do not fully account for.
%
% Experimental results demonstrate that POEM significantly improves the accuracy of localized transformations while maintaining semantic consistency, setting a new standard for precise object-level image editing. \dimpp{hmmm, too vague?}
Experimental results demonstrate that POEM achieves significantly higher edit fidelity compared to existing text-based editing approaches while requiring no additional user annotations, unlike interaction-based methods.

Our contributions are two-fold: (a) we introduce a plug-and-play reasoning block that interprets user edit instructions with high numerical precision, generating accurate object masks and transformation matrices that enhance layout modifications and mask-guided diffusion editing; (b) we present VOCEdits, a novel dataset for evaluating precise object-level edits, establishing a comprehensive benchmark for detection, transformation, and synthesis tasks.
%
%This work lays a foundation for future research in multimodal image editing and advances the controllability of diffusion models. We will release the dataset, the code, and the models upon acceptance of the paper.

%% file: 2_related_work.tex
\section{Related Work}
\label{sec:related}

% \dimpp{I tried to make this section smaller. I turned everything into paragraphs instead of subsections. Still some parts are a bit long, but we can check it later.}

% \subsection{Controlling Diffusion Models}
% \label{sec:2.1}

\mypar{Controlling Diffusion Models}.
Stable Diffusion \cite{podellSDXLImprovingLatent2023,rombachHighResolutionImageSynthesis2022} has become a leading model for high-resolution image generation. Recent efforts have explored various approaches for controlling such models, broadly categorized into guidance~\cite{hoClassifierFreeDiffusionGuidance2022}, fine-tuning~\cite{ruizDreamBoothFineTuning2023}, textual inversion~\cite{galImageWorthOne2022}, and attention control~\cite{hertzPrompttoPromptImageEditing2022}. Guidance methods~\cite{hoClassifierFreeDiffusionGuidance2022} steer the generation process using auxiliary signals, such as class labels, or text.
Fine-tuning~\cite{ruizDreamBoothFineTuning2023} modifies model weights to associate edit prompts with example images. Textual Inversion~\cite{mokadyNulltextInversionEditing2023} optimizes concepts within the text encoder's embedding space.
Finally, attention control~\cite{hertzPrompttoPromptImageEditing2022} modifies spatial attention maps within diffusion layers to influence layout and geometry, enabling precise structural preservation while allowing targeted contextual edits.  

% \subsection{Image Editing with Diffusion Models}
% \label{sec:2.2}

% With improved control over diffusion models, the focus has increasingly shifted from pure image generation to refined image editing, where modifications must not only align with user intent but also preserve the original image structure. Recent image editing methods can be compared based on different input modalities.

\mypar{Text-to-image editing} extends the foundational image-guided generation approaches. Early methods~\cite{hertzPrompttoPromptImageEditing2022,mengSDEditGuidedImage2022} edit images by manipulating cross-attention maps. Imagic~\cite{kawarImagicTextBasedReal2023} finetunes the model at inference time to directly match text prompts with visual outputs while striving to preserve the image's style and structure. In contrast, InstructPix2Pix (IP2P)~\cite{brooksInstructPix2PixLearningFollow2023} eliminates inference-time fine-tuning by using classifier-free guidance to condition on the source image and text prompt.
While IP2P enables global edits, it often over-modifies images, prompting further research~\cite{guoFocusYourInstruction2024,mirzaeiWatchYourSteps2023,sheyninEmuEditPrecise2023} into more localized edits.

\mypar{Editing with image interaction inputs} offers control beyond text-based methods. These approaches require users to provide additional guidance through masks~\cite{lugmayrRePaintInpaintingUsing2022,nicholGLIDEPhotorealisticImage2022,parkShapeGuidedDiffusionInsideOutside2024,xieSmartBrushTextShape2023}, or point dragging~\cite{cuiStableDragStableDragging2024,gengMotionGuidanceDiffusionBased2024,mouDiffEditorBoostingAccuracy2024}
and use them to optimize latent codes more precisely. 
% %
% Image-to-image approaches like 
% ControlNet \cite{} 
% use auxiliary inputs like pose sketches for a more structured control. 
% %
% Exemplar-based methods~\cite{srivastavaReEditMultimodalExemplarBased2024,chenZeroshotImageEditing2024,yangPaintExampleExemplarbased2023} rely on reference images for direct appearance edits, enabling intuitive and adaptable modifications.

\begin{figure}[t]
    \centering
    \includegraphics[width=\linewidth]{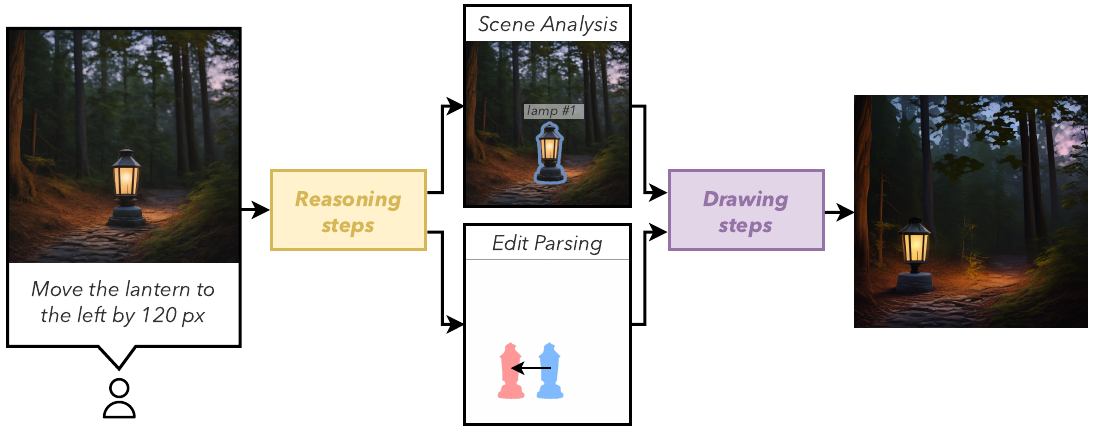}
    \caption{\textbf{Overview of our approach.} An image and a user edit prompt are fed into the reasoning stage, where we analyze the scene and extract object-level masks and precise transformation parameters for appearance and shape edits. 
    %SAM and LLMs refine object-level binary masks and transformation parameters, enabling higher precision edits. 
    During the editing stage, we apply these edits during inference without any additional training or fine-tuning.}
    \label{fig:2}
\end{figure}

\mypar{Multimodal Large Language Models (MLLMs)}
enhance image editing workflows~\cite{brooksInstructPix2PixLearningFollow2023} by interpreting context-aware user instructions~\cite{huangSmartEditExploringComplex2024}. They resolve ambiguities, capture the underlying user intents~\cite{yuPromptFixYouPrompt2024}, and are adept at handling long and detailed edit prompts~\cite{liuImprovingLongTextAlignment2024}. Another line of work focuses on layout composition and canvas-based image editing by integrating MLLMs and LLMs to enforce robust object-attribute binding and multi-subject descriptions~\cite{fengLayoutGPTCompositionalVisual2023,fengRanniTamingTexttoImage2024,wuSelfcorrectingLLMcontrolledDiffusion2023,zengSceneComposerAnyLevelSemantic2023}. 
For example, Ranni~\cite{fengRanniTamingTexttoImage2024}  
enhances textual controllability using a semantic panel, while
SceneComposer~\cite{zengSceneComposerAnyLevelSemantic2023} enables synthesis from textual descriptions to precise 2D semantic layouts. LayoutGPT \cite{fengLayoutGPTCompositionalVisual2023} acts as a visual planner for generating layouts from text, and SLD \cite{wuSelfcorrectingLLMcontrolledDiffusion2023} iteratively refines images by employing LLMs to analyze the prompt and improved alignment. MLLMs also serve as orchestrators, decomposing complex edits into subtask tree, selecting tools, and coordinating their use~\cite{wangGenArtistMultimodalLLM2024,weiOmniEditBuildingImage2024}. 
%They analyze the task at hand, select the most appropriate tools, and coordinate their use, thereby seamlessly integrating generation and editing processes.
%
Unlike previous methods, we utilize MLLMs to conduct visual reasoning based on edit instructions and source images, focusing on their strengths in numerical proficiency. This enables precise control, such as affine transformation parameters applied to object-level shapes. 

%% file: 3_method.tex
\section{Method}
\label{sec:3}

\begin{figure}[t]
    \centering
    \includegraphics[width=\linewidth]{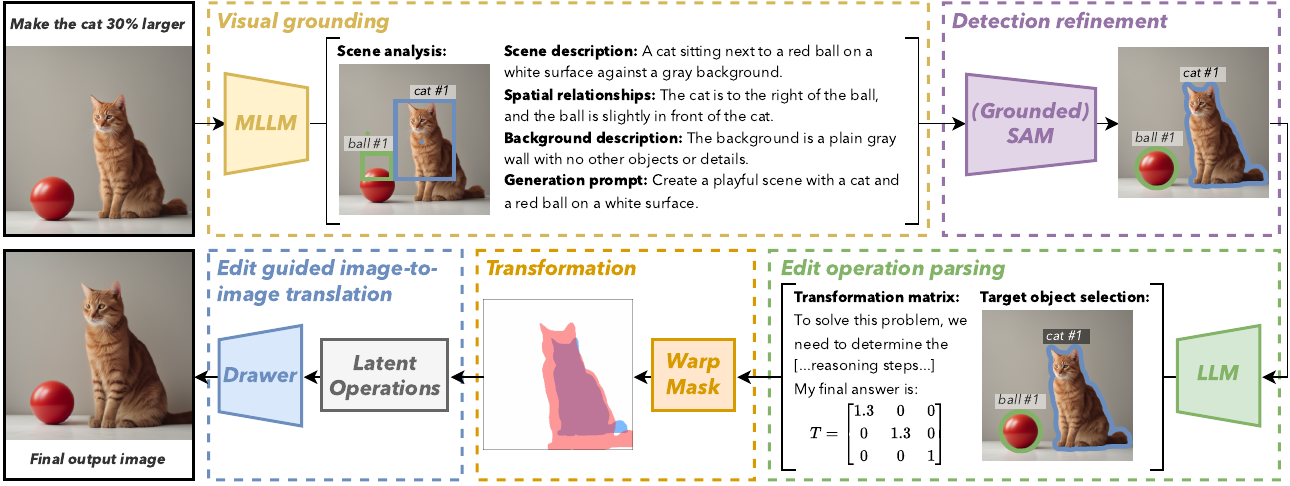}
    \caption{\textbf{Detailed pipeline of POEM.} Given an image and an edit prompt, we first use an MLLM to analyze the scene and identify objects. Then, we refine the detections and enhance object masks using Grounded SAM. Next, we use a text-based LLM to predict the transformation matrix of the initial segmentation mask. Finally, we perform an image-to-image translation guided by the previous steps to generate the edited image. This structured pipeline enables precise object-level editing with high visual fidelity while preserving spatial and visual coherence.}
    \label{fig:3}
\end{figure}

Given an input image $I$ and a textual edit instruction $P$, our goal is to generate a modified image $\hat{I}$ that reflects precise object-level transformations specified in $P$. To do that, we leverage MLLM-driven reasoning to eliminate the need for additional user interaction.
We propose POEM (Precise Object-level Editing via MLLM Control), an approach designed for high-precision object-level image editing.
POEM decouples the visual reasoning from the image editing (drawing) to achieve fine-grained object transformations (Fig.~\ref{fig:2}).

POEM consists of five steps (Fig.~\ref{fig:3}): 
(a) Visual Grounding: the input image and the edit prompt are fed into an MLLM 
that is instructed to analyze the scene and identify and detect all objects;
(b) Detection Refinement: we refine the object detection output from the MLLM to obtain more accurate object segmentation masks;
(c) Edit Operation Parsing: we use an LLM that is instructed to select the target object and compute the transformation matrix;
(d) Transformation: we apply the transformation to the segmented object to obtain the edited mask and
(e) Edit Guided Image-to-Image Translation: given the initial input image and the masks of the target object before and after the transformation, we generate the final modified image while preserving spatial and visual coherence.

% \subsection{Visual Grounding}
% \label{sec:3.1}

\mypar{Visual Grounding.}
In this step, we deploy an MLLM that takes as input the image $I$ and the prompt $P$. 
Using zero-shot prompting, we leverage the model's visual capabilities to analyze the scene and detect all objects in the image. The MLLM is directly instructed to perform object detection and detect all objects $N$ that appears in the image. For each detected object $i \in N$, we ask the MLLM to output the detected bounding box $b_i$, a segmentation point on the object $s_i$, the object class $c_i$, and a unique object ID $k_i$. 

%% reviwer 2 comments (shortene)
Additionally, we instruct the MLLM to analyze the image $I$ and user prompt $P$, generating four structured descriptions: the scene ($S$), spatial relationships ($R$), background prompt ($P_{bg}$), and generation prompt ($P_g$). These are not direct captions but targeted summaries capturing (1) global layout ($S$), (2) object relationships ($R$), (3) background appearance and context ($P_{bg}$), and (4) overall generation intent ($P_g$). $S$ and $R$ support Edit Operation Parsing to estimate the transformation matrix, while $P_{bg}$ and $P_g$ guide the Drawer step to maintain background consistency and apply object-specific edits.

%% Marco -- Reviewer 2
% Additionally, we instruct the MLLM to analyze the image $I$ and the user prompt $P$, and generate four textual descriptions—namely the scene  ($S$), the spatial relationship  ($R$), the background-prompt  ($P_{bg}$), and a generation-prompt  ($P_g$). These are not direct image captions, but structured descriptions that capture (1) the global scene layout ($S$), (2) spatial relationships between relevant objects ($R$), (3) background appearance and context ($P_{bg}$), and (4) overall generation intent ($P_g$). Each of these descriptions serves a targeted role within our pipeline: $S$ and $R$ are used during the Edit Operation Parsing step to help the LLM estimate the transformation matrix, while $P_{bg}$ and $P_g$ are used at alst in the drawer step to guide the diffusion-based editor to preserve background consistency and apply object-focused edits. 

%% original Submission
% Additionally, we instruct the MLLM to analyze $I$ and $P$ and provide four descriptive prompts --$S$,$R$,$P_{bg}$, $P_g$-- capturing details about the scene, the spatial relationship of the objects, the background, and the image generation process, respectively.

% \subsection{Detection Refinement}
% \label{sec:3.2}

\mypar{Detection Refinement.}
Off-the-shelf MLLMs often struggle to produce precise object-bounding boxes when performing a visual grounding task~\cite{hanoona2023GLaMM}. For this reason, in this step, we refine the detection output from the previous step, and for each detected object $i$, we obtain a segmentation mask $m_i$ and a refined bounding box $b_i'$.
Without loss of generality, we use Grounded-SAM~\cite{renGroundedSAMAssembling2024}, and we prompt it with the predicted object classes $c_i$ from the previous step. Grounded-SAM combines Grounding DINO~\cite{liu2023grounding} and Segment Anything Model (SAM)~\cite{raviSAM2Segment2024} to perform an open-set detection and segmentation with text prompts even for objects outside predefined
categories. 

% \subsection{Edit Operation Parsing}
% \label{sec:3.3}

\mypar{Edit Operation Parsing.}
Given the prompt $P$ and the set of the refined bounding boxes $B'=\{b_i'|i \in N\}$, the goal of this step is to extract a transformation matrix $T$ and identify the ID $k$ of the 
target object. 
Given only the prompt $P$, the MLLM from the first step struggles to directly infer $T$ in a single step due to its lack of explicit scene information. For instance, if \texttt{P = 'make the cat 100px wide'}, the required transformation depends on the cat’s initial dimensions in the image. If the cat is 50px wide, the scaling factor in $T$ should be $2$, whereas if the initial width is 25px, the scaling factor in $T$ should be $4$.

To address this, we use a text-based LLM optimized for mathematical reasoning to compute the transformation parameters. This separation allows for a more accurate estimation of scale, rotation, and translation transformations by explicitly incorporating object size information into the reasoning process. We use the input prompt $P$, the descriptive prompts $S$ and $R$, and the coordinates of the detections $B'$, and we directly instruct the LLM to predict the unique ID of the target object $i*$ and a $3x3$ transformation matrix $T$ given by:

\begin{equation}
T = \begin{bmatrix} a_{11} & a_{12} & a_{13} \\ a_{21} & a_{22} & a_{23} \\ 0 & 0 & 1 \end{bmatrix}
\end{equation}

\noindent To ensure precise parsing, we employ a structured format where LLM matrices and object IDs are enclosed between the unique tokens \texttt{<MSTART>},\texttt{\texttt{<MEND>}}, \texttt{<ISTART>}, and \texttt{<IEND>}. A regex-based parser extracts numerical values enclosed within the matrix tokens, ensuring the retrieval of transformation parameters.

% \subsection{Transformation}
% \label{sec:3.4}

\mypar{Transformation.}
In this step, we select the segmentation mask $m_{i*}$ corresponding to the selected id $i*$. Then, we perform image wrapping using $T$ on the binary mask $m_{i*}$ to generate the transformed mask $\hat{m}_{i*}$.

% \subsection{Edit Guided Image-to-Image Translation}
% \label{sec:3.5}

\mypar{Edit Guided Image-to-Image Translation.}
In this step, we use the masks $m_{i*}$ and $\hat{m}_{i*}$ of the target object, and the descriptive prompts $P_{bg}$ and $P_g$ from the first step to perform the image synthesis and generate the final input image $\hat{I}$. We apply these edits during the inference of pre-trained diffusion models without additional training or fine-tuning.
Inspired by~\cite{wuSelfcorrectingLLMcontrolledDiffusion2023}, we perform object-level shape manipulations in the latent space of diffusion models~\cite{rombachHighResolutionImageSynthesis2022}. We use the region of the mask $\hat{m}_{i*}$ to define the area of interest, which is processed through backward diffusion to obtain its latent representation \(z_{\text{repos}}\). The region of the initial maks $m_{i*}$ is reinitialized with Gaussian noise \(\mathcal{N}(0, I)\), and the new latent is blended into the image latent \(z\) as:

\begin{equation}
z_{\text{new}} = z \odot (1 - M_j) + z_{\text{repos}} \odot \hat{M}_j + \mathcal{N}(0, I) \odot M_j.
\end{equation}

\noindent A forward diffusion process refines the image, enhancing realism and coherence in edited and surrounding regions.

%% file: 4_experiments.tex
\section{Experiments}
\label{sec:4}

\begin{table}[t]\centering
    \renewcommand{\arraystretch}{1.2} % Increase row spacing
    \setlength{\tabcolsep}{6pt} % Adjust column spacing
    \caption{\textbf{Evaluation on VOCEdits.} Methods are grouped according to different steps of our pipeline, as described in the paper. For each step, we report the Intersection over Union (IoU) (\%) between the sets indicated in the right part of the table. The final section of the table presents results from other state-of-the-art image editing methods.}
    \label{tab:1-main-quantitative}

\resizebox{\linewidth}{!}{
\begin{tabular}{l|cccccccc}
\toprule
\rowcolor{gray!15} 
\textbf{Method} & Move & Scale & Flip & Shear & Rotate & Reason & Mix & Avg. \\
\midrule
\rowcolor{sec31_yellow!25} \multicolumn{1}{l|}{\textbf{Visual Grounding}} & \multicolumn{8}{c}{Estimated bounding box in the input image vs. GT} \\
InternVL-8B~\cite{chen2024internvl}       & 15.8 & 19.8  & 9.4  & 23.3 & 13.4 & 27.3 & 24.3 & 17.4 \\
InternVL-72B~\cite{chen2024internvl}   & 46.3 & 47.4  & 43.9 & 49.6 & 49.6 & 43.4 & 46.8 & 47.1 \\
QwenVL-7B~\cite{qwenQwen25TechnicalReport2025}       & 54.8 & \textbf{57.8} & \textbf{54.0} & \textbf{55.1} & 54.0 & 37.8 & \textbf{50.1} & \textbf{55.5} \\
QwenVL-72B~\cite{qwenQwen25TechnicalReport2025}   & \textbf{55.1} & 56.4  & 53.5 & 54.2 & \textbf{54.6} & \textbf{45.1} & 40.7 & 54.8 \\
\midrule
\rowcolor{sec32_violet!25} \multicolumn{1}{l|}{\textbf{Detection Refinement}} & \multicolumn{8}{c}{Estimated segmentation mask in the input image vs. GT} \\
QwenVL-7B~\cite{qwenQwen25TechnicalReport2025}  + SAM~\cite{raviSAM2Segment2024}                      & 22.5 & 34.1 & 22.5 & 20.0  & 24.0 & 31.4 & 27.0 & 27.3 \\
QwenVL-7B~\cite{qwenQwen25TechnicalReport2025}  + G-SAM~\cite{renGroundedSAMAssembling2024}                     & \textbf{82.6} & \textbf{86.0} & \textbf{81.3} & \textbf{81.0} & \textbf{88.5} & \textbf{51.3} & \textbf{81.3} & \textbf{84.2} \\
\midrule
\rowcolor{sec33_green!25} \multicolumn{1}{l|}{\textbf{Edit Operation Parsing \& Transformation}} & \multicolumn{8}{c}{Transformed segmentation mask vs. GT} \\
(QwenVL-7B~\cite{qwenQwen25TechnicalReport2025}  + G-SAM~\cite{renGroundedSAMAssembling2024}) + DeepSeek~\cite{deepseek-aiDeepSeekV3TechnicalReport2024}       & 20.6 & 26.4 & 30.7 & 38.6 & 28.1 & 2.5 & 21.7 & 25.3 \\
(QwenVL-7B~\cite{qwenQwen25TechnicalReport2025}  + G-SAM~\cite{renGroundedSAMAssembling2024}) + QwenM~\cite{yangQwen25MathTechnicalReport2024}          & \textbf{42.0} & \textbf{50.3}  & \textbf{58.0} & \textbf{80.3}  & \textbf{52.5} & \textbf{9.5} & \textbf{37.7} & \textbf{49.2} \\
\midrule
Oracle Mask + DeepSeek~\cite{deepseek-aiDeepSeekV3TechnicalReport2024}            & 23.1 & 30.6 & 40.0  & 56.1 & 30.2  & 2.1 & 21.4 & 29.5 \\
Oracle Mask + QwenM~\cite{yangQwen25MathTechnicalReport2024}                & \textbf{47.0} & \textbf{56.0} & \textbf{74.1} & \textbf{98.6} & \textbf{56.2} & \textbf{17.5} & \textbf{38.8} & \textbf{55.6} \\
\midrule
\rowcolor{sec35_blue!25} \multicolumn{1}{l|}{\textbf{Edit Guided Image-to-Image Translation}} & \multicolumn{8}{c}{Detected segmentation mask in the output image vs. GT} \\
(QwenVL-7B~\cite{qwenQwen25TechnicalReport2025}  + G-SAM~\cite{renGroundedSAMAssembling2024}  + QwenM~\cite{yangQwen25MathTechnicalReport2024}) + SLD~\cite{wuSelfcorrectingLLMcontrolledDiffusion2023}            & \textbf{32.6} & \textbf{39.7}  & 39.1 & 54.5 & 43.5 & 24.9 & 37.6 & \textbf{38.4} \\
(QwenVL-7B~\cite{qwenQwen25TechnicalReport2025}  + G-SAM~\cite{renGroundedSAMAssembling2024}  + QwenM~\cite{yangQwen25MathTechnicalReport2024}) + SLD~\cite{wuSelfcorrectingLLMcontrolledDiffusion2023}  +~\cite{podellSDXLImprovingLatent2023}   & 31.6 & 39.4 & 37.4 & 53.3 & 42.2 & 24.7 & 36.1 & 37.6 \\
\midrule
IP2P~\cite{brooksInstructPix2PixLearningFollow2023}   & 27.4 & 32.9 & 41.9 & \textbf{72.3} & 38.3 & 8.4 & 33.8 & 34.3 \\
TurboEdit~\cite{deutch2024turboedit}   & 27.1 & 30.9 & 46.4 & 53.6 & 43.9 & 21.8 & 39.7 & 33.8 \\
LEDITS++~\cite{brackLEDITSLimitlessImage2024}   & 27.4 & 31.8 & \textbf{52.7} & 56.2 & \textbf{44.1} & \textbf{29.9} & \textbf{39.8} & 35.0 \\
\bottomrule
\end{tabular}
}
\end{table}

This section presents our experimental results. We introduce VOCEdits, a novel dataset to ensure rigorous evaluation of precise object-level edits in Sec.~\ref{sec:4.1}. 
In Sec.~\ref{sec:4.2}-\ref{sec:4.5}, we systematically explore different design choices for each step of our pipeline and evaluate their impact. We also present a qualitative comparison between POEM and state-of-the-art image editing approaches ~\cite{brackLEDITSLimitlessImage2024,brooksInstructPix2PixLearningFollow2023,deutch2024turboedit}, while in Sec.~\ref{sec:4.7}, we discuss the limitations of our approach.

\subsection{VOCEdits Dataset}
\label{sec:4.1}
We present VOCEdits, a dataset for evaluating fine-grained object-level image editing involving affine transformations: flip, scale, rotation, translation, and shear. It is built upon PASCAL VOC 2012~\cite{everingham2015pascal} for its high-quality instance segmentation masks, enabling precise object-centric evaluation on real-world images. We augment PASCAL VOC images with instructional prompts, ground-truth transformations, and object masks before and after editing. We use images from the PASCAL VOC 2012 trainval segmentation set, containing 2913 images and 6929 object instances. We filter out images with multiple instances of the same class, truncated objects, extreme object sizes, or masks extending beyond image boundaries, resulting in 505 unique images.

% We present the VOCEdits, a dataset designed to evaluate fine-grained object-level image editing involving affine transformations, such as flip, scale, rotation, translation, and shear. The dataset is built upon PASCAL VOC 2012~\cite{everingham2015pascal}  due to its high-quality instance segmentation masks, enabling precise object-centric evaluations of real-world images. We augment the images of PASCAL VOC with instructional edit prompts, ground-truth transformations, and precise ground-truth object masks for before-and-after transformations.
% We obtain the images from the trainval set of PASCAL VOC 2012(Segmentation Task), which contains 2913 images and 6929 object instances.
% We first filter out images containing multiple instances of the same class, truncated objects, excessively small or large objects, or segmentation parts extending beyond image boundaries. This leads to a set of 505 unique images. 

To generate human-like edit instructions, we utilize GPT-4o-mini~\cite{achiam2023gpt} and instruct it to paraphrase default instructions, leading to diverse descriptions. The ground-truth object segmentation masks from PASCAL VOC  are then transformed via open-cv transformation for exact computation. Each image from the final set undergoes two randomly selected transformations, with three corresponding paraphrased prompts, yielding a total of 3030 unique samples.

\begin{figure}[t]
    \centering
    % \begin{tabular}{@{} 
    %   >{\centering\arraybackslash}m{0.33\linewidth} @{} 
    %   >{\centering\arraybackslash}m{0.33\linewidth} @{} 
    %   >{\centering\arraybackslash}m{0.33\linewidth} @{} 
    %   }
    %     % Header row (centered)
    %     \scriptsize (a) Object class
    %     & \scriptsize (b) Transformation type
    %     & \scriptsize (c) Transformation difficulty \\%[3pt]
        
    %     % Example content rows
    %     \includegraphics[width=\linewidth,page=2]{figures/DatasetStatistics.pdf}
    %     & \includegraphics[width=\linewidth,page=1]{figures/DatasetStatistics.pdf}
    %     & \includegraphics[width=\linewidth,page=4]{figures/DatasetStatistics.pdf} \\%[3pt]
    % \end{tabular}
    
\includegraphics[width=\linewidth]{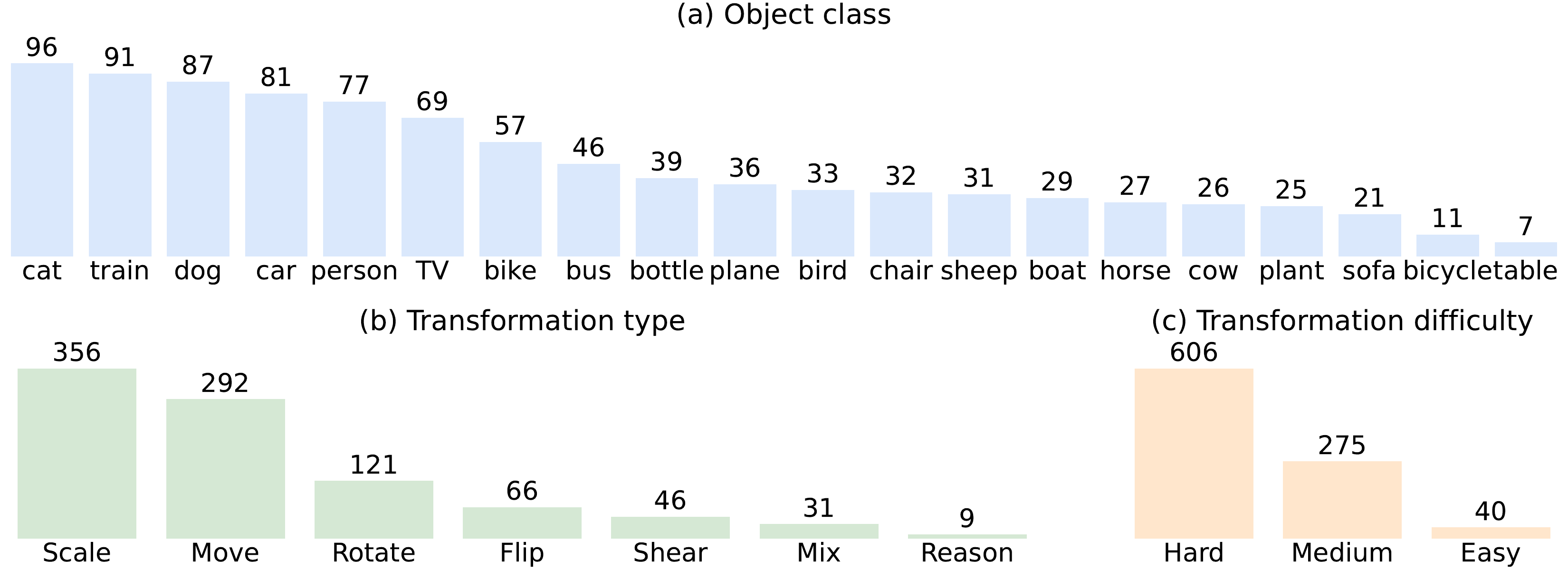}
    
% \caption{\textbf{VOCEdits evaluation subset statistics.} We show (a) the distribution of object classes, (b) the distribution of the transformation types, and (c) the distribution of transformation difficulty levels. The dataset contains a diverse range of objects, with scaling and translation being the most common transformations.}
% Most transformations fall into the ``hard" category, highlighting the complexity of the evaluation set.
%\dimpp{change compose to mix and reasoning to reason}
\caption{\textbf{VOCEdits evaluation subset statistics.} Distributions of (a) object classes, (b) transformation types, and (c) transformation difficulty levels.}
    \label{fig:3-datasetstatistics}
    \vspace{-2mm}
\end{figure}

Our pipeline processes all 3030 samples but applies an additional refinement step, excluding cases with more than five foreground objects per image. This restriction is imposed due to the limitations of~\cite{wuSelfcorrectingLLMcontrolledDiffusion2023} in handling excessive object occlusions and intersecting boxes. After this filtering, a final set of 193 images and 921 samples is retained for evaluation. Unless stated otherwise, we will use this set to evaluate our pipeline for the remainder of this section. A comprehensive summary of these results is provided in Tab.~\ref{tab:1-main-quantitative}.
Fig.~\ref{fig:3-datasetstatistics} provides detailed statistics on transformation distribution and object categories of the final set.

\begin{figure}[t]
    \centering
    
% CHERRY PICKED - OLD
% \begin{tabular}{@{} 
%   >{\centering\arraybackslash}m{0.04\linewidth} @{} 
%   >{\centering\arraybackslash}m{0.225\linewidth} @{} 
%   >{\centering\arraybackslash}m{0.225\linewidth} @{} 
%   >{\centering\arraybackslash}m{0.02\linewidth} @{} 
%   >{\centering\arraybackslash}m{0.04\linewidth} @{} 
%   >{\centering\arraybackslash}m{0.225\linewidth} @{} 
%   >{\centering\arraybackslash}m{0.225\linewidth} @{}
%   }
%     % Header row (centered)
%     & \scriptsize Translate
%     & \scriptsize Scale
%     & 
%     & 
%     & \scriptsize Translate
%     & \scriptsize Scale \\%[3pt]
    
%     % Example content rows
%     \rotatebox{90}{\scriptsize IP2P} 
%     & \includegraphics[width=\linewidth]{figures/Comparison/ip2p_move.png}
%     & \includegraphics[width=\linewidth]{figures/Comparison/ip2p_resize.png}
%     & 
%     & \rotatebox{90}{\scriptsize TurboEdit} 
%     & \includegraphics[width=\linewidth]{figures/Comparison/turboedit_move.png}
%     & \includegraphics[width=\linewidth]{figures/Comparison/turboedit_resize.png} \\%[3pt]

%     \rotatebox{90}{\scriptsize LEDITS++} 
%     & \includegraphics[width=\linewidth]{figures/Comparison/ledits_move.png}
%     & \includegraphics[width=\linewidth]{figures/Comparison/ledits_resize.png}
%     & 
%     & \rotatebox{90}{\scriptsize POEM (Ours)} 
%     & \includegraphics[width=\linewidth]{figures/Comparison/ours_afterrefine_move.png}
%     & \includegraphics[width=\linewidth]{figures/Comparison/ours_afterrefine_resize.png} \\%[3pt]
% \end{tabular}

% Different operations
\begin{tabular}{@{} 
>{\centering\arraybackslash}m{0.04\linewidth} @{\hspace{0.015\linewidth}}
>{\centering\arraybackslash}m{0.145\linewidth} @{\hspace{0.015\linewidth}}
>{\centering\arraybackslash}m{0.145\linewidth} @{\hspace{0.015\linewidth}}
>{\centering\arraybackslash}m{0.145\linewidth} @{\hspace{0.015\linewidth}}
>{\centering\arraybackslash}m{0.145\linewidth} @{\hspace{0.015\linewidth}}
>{\centering\arraybackslash}m{0.145\linewidth} @{\hspace{0.015\linewidth}}
>{\centering\arraybackslash}m{0.145\linewidth}
}
  
    % Header row (centered)
    & \scriptsize Scale x0.56
    & \scriptsize Move left 150px and make it red
    & \scriptsize Scale only vertical 200px
    & \scriptsize Make the sword gold
    & \scriptsize Scale x2, move left 150px
    & \scriptsize Move left 90px, make it blue\\%[3pt]
    
    % Example content rows
    \rotatebox{90}{\scriptsize Input images} 
    & \includegraphics[width=\linewidth]{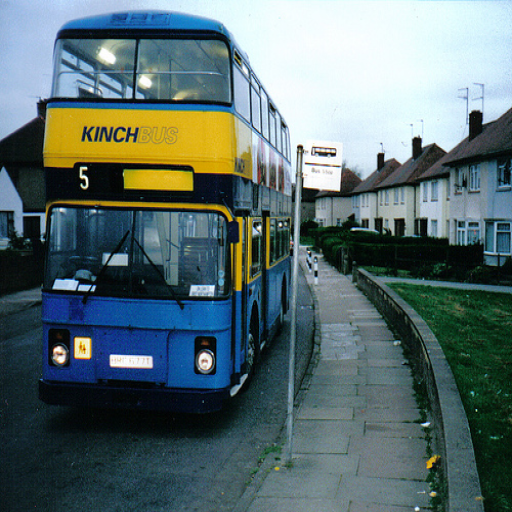}
    & \includegraphics[width=\linewidth]{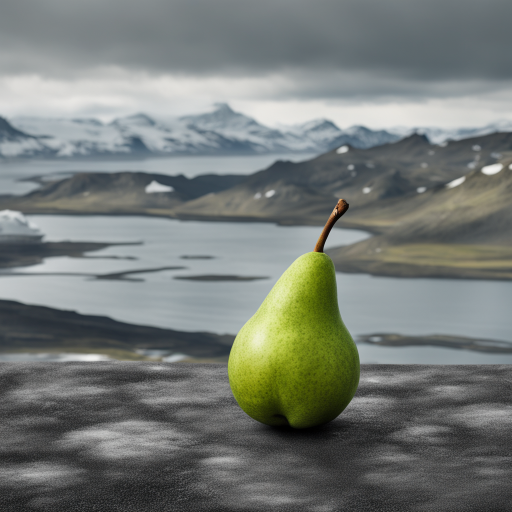}
    & \includegraphics[width=\linewidth]{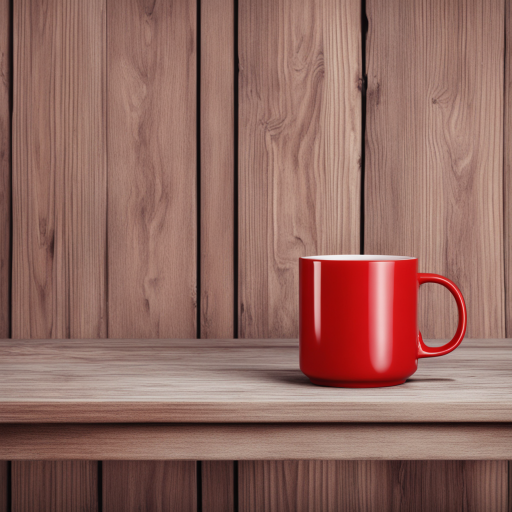}
    & \includegraphics[width=\linewidth]{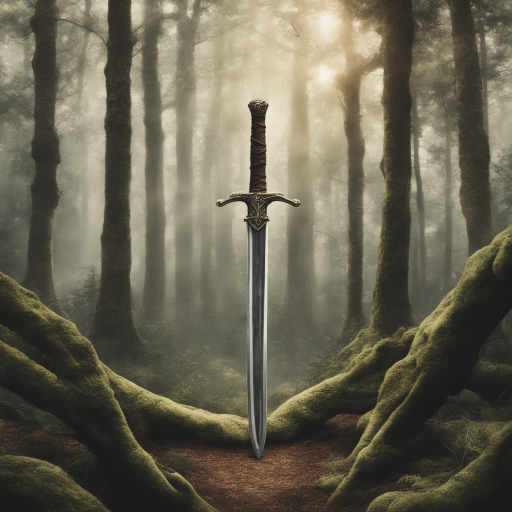}
    & \includegraphics[width=\linewidth]{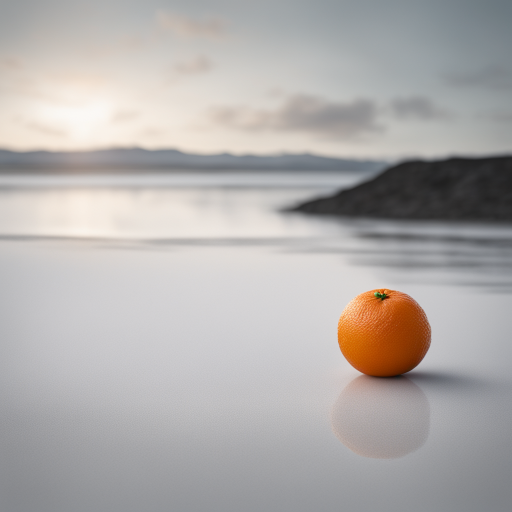}
    & \includegraphics[width=\linewidth]{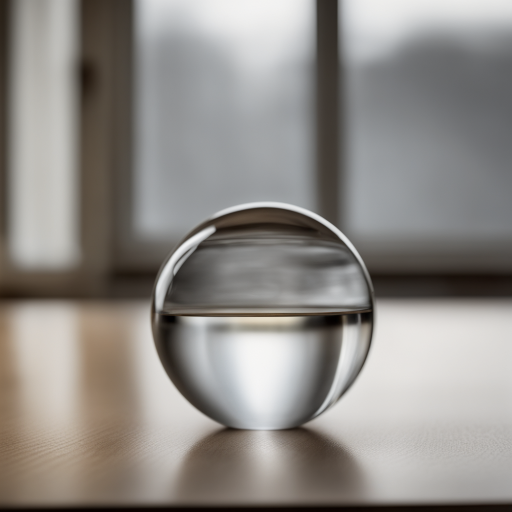}\\

    \rotatebox{90}{\scriptsize IP2P} 
    & \includegraphics[width=\linewidth]{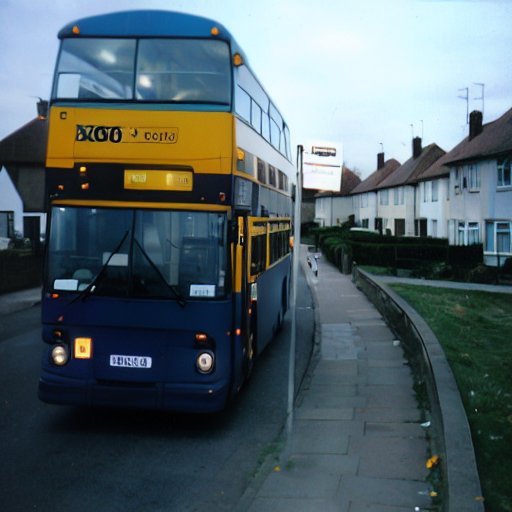}
    & \includegraphics[width=\linewidth]{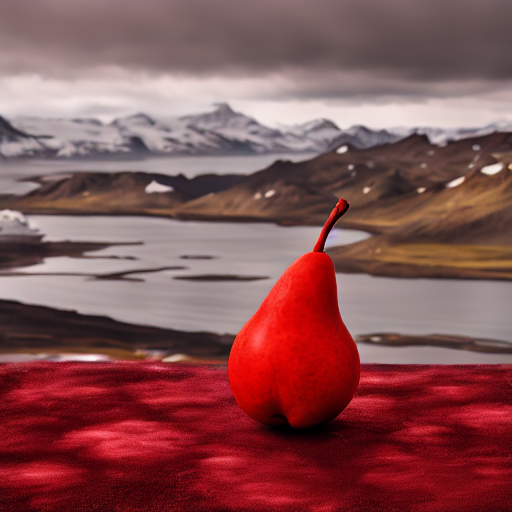}
    & \includegraphics[width=\linewidth]{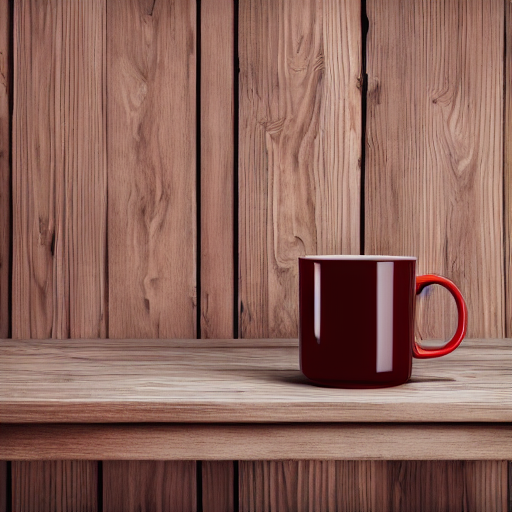}
    & \includegraphics[width=\linewidth]{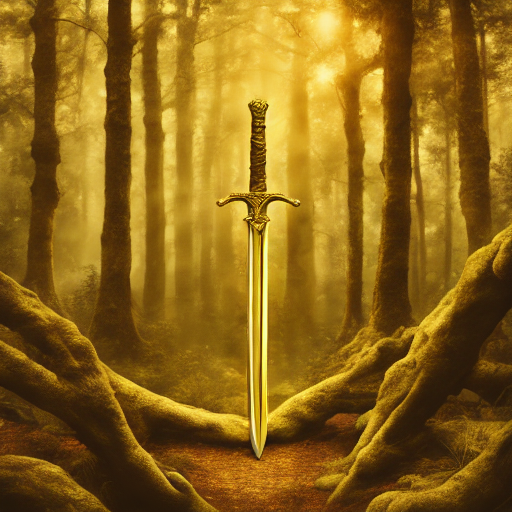}
    & \includegraphics[width=\linewidth]{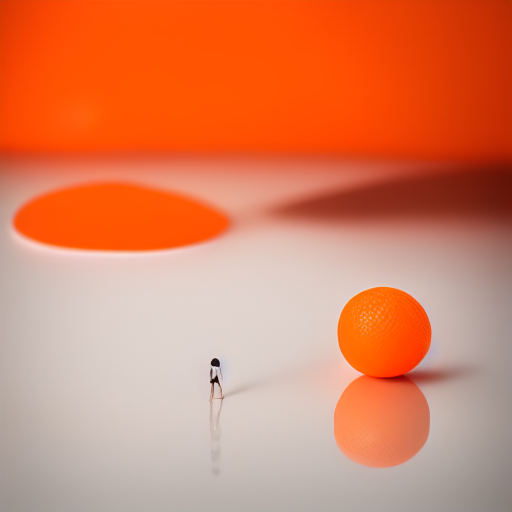}
    & \includegraphics[width=\linewidth]{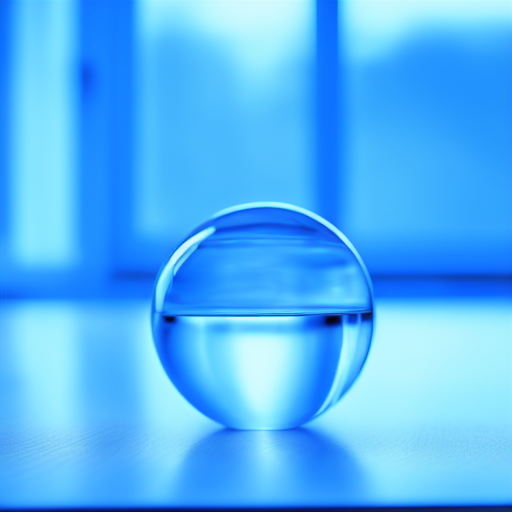}\\

    \rotatebox{90}{\scriptsize TurboEdit} 
    & \includegraphics[width=\linewidth]{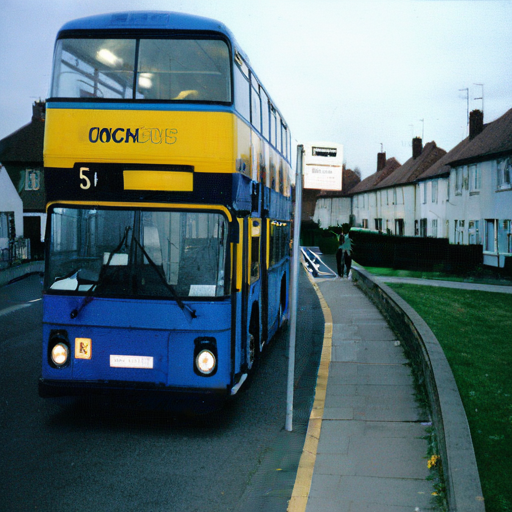}
    & \includegraphics[width=\linewidth]{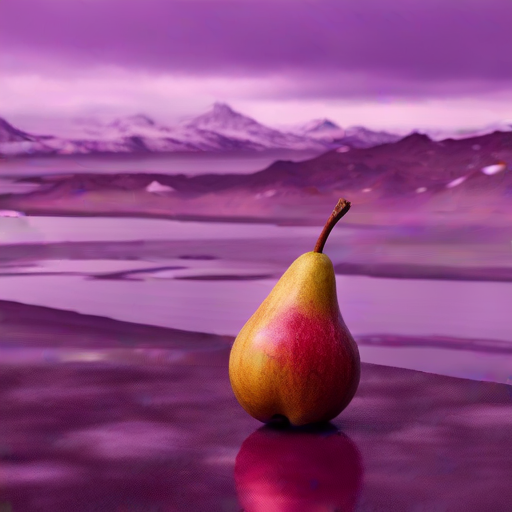}
    & \includegraphics[width=\linewidth]{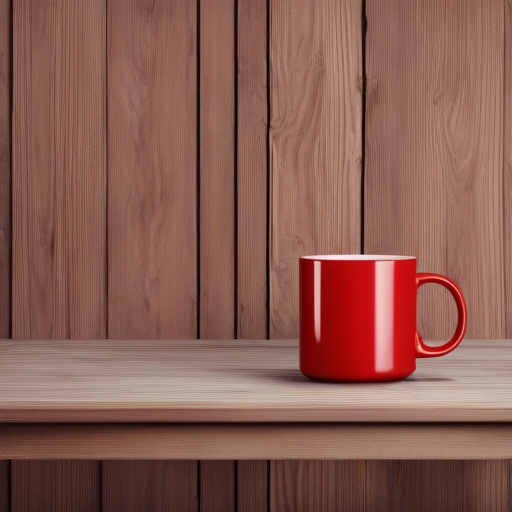}
    & \includegraphics[width=\linewidth]{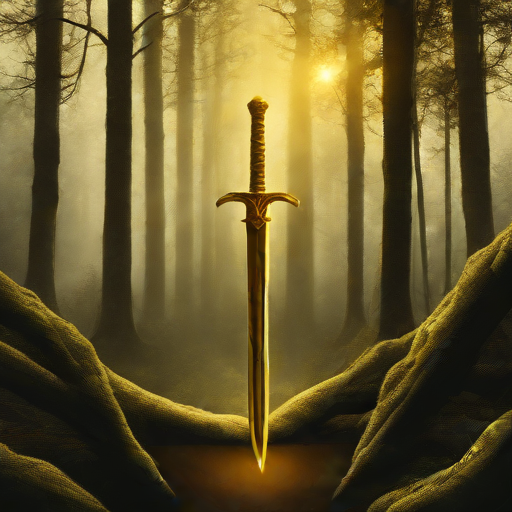}
    & \includegraphics[width=\linewidth]{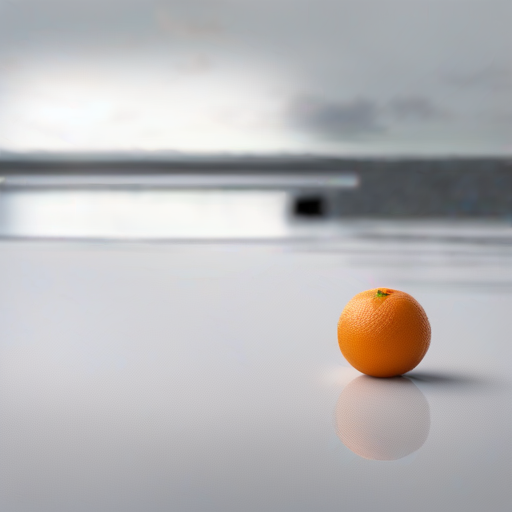}
    & \includegraphics[width=\linewidth]{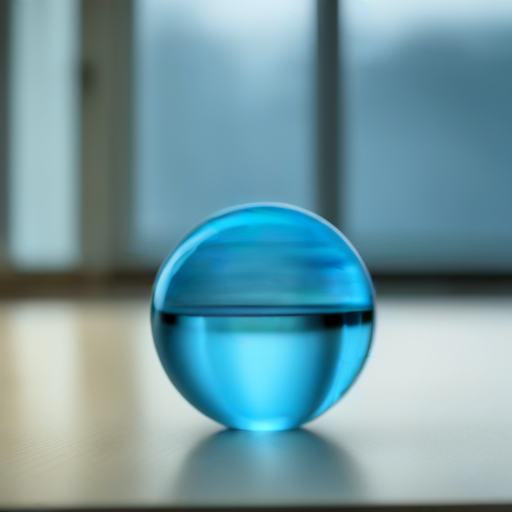}\\%[3pt]

    \rotatebox{90}{\scriptsize LEDITS++} 
    & \includegraphics[width=\linewidth]{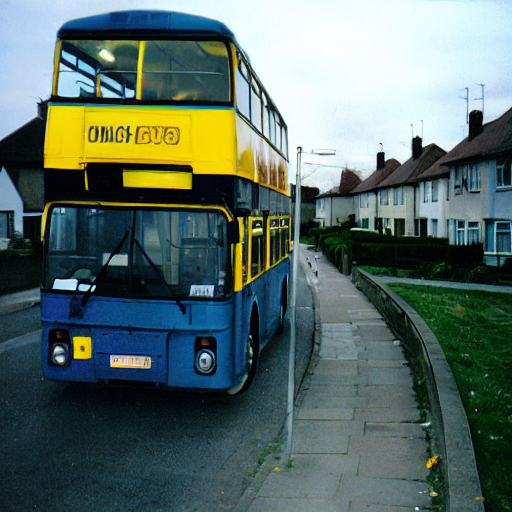}
    & \includegraphics[width=\linewidth]{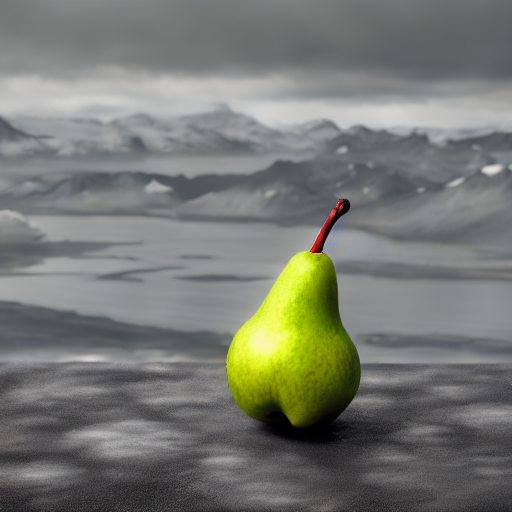}
    & \includegraphics[width=\linewidth]{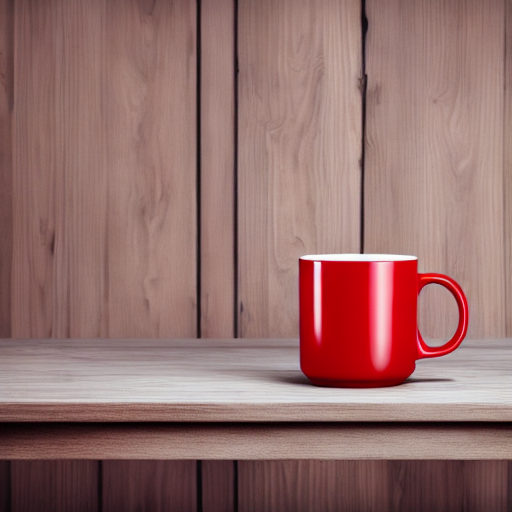}
    & \includegraphics[width=\linewidth]{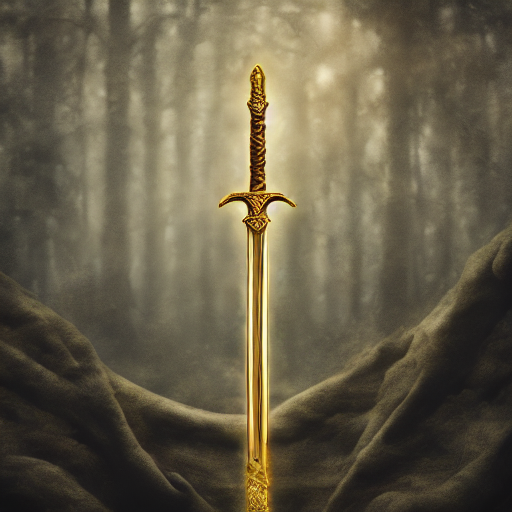}
    & \includegraphics[width=\linewidth]{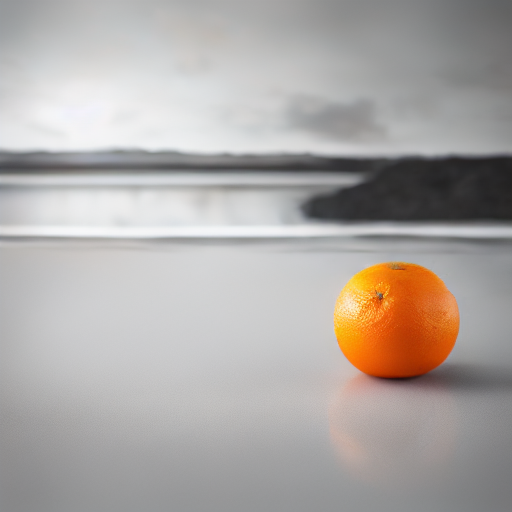}
    & \includegraphics[width=\linewidth]{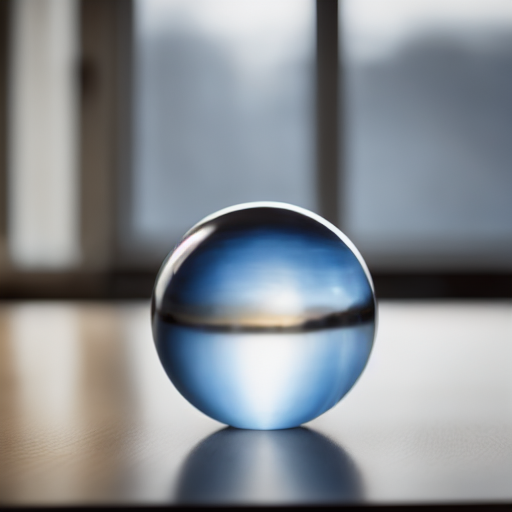}\\
    
    \rotatebox{90}{\scriptsize POEM (Ours)} 
    & \includegraphics[width=\linewidth]{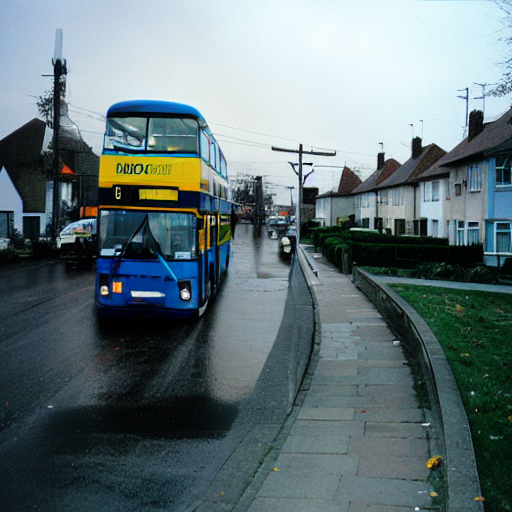}
    & \includegraphics[width=\linewidth]{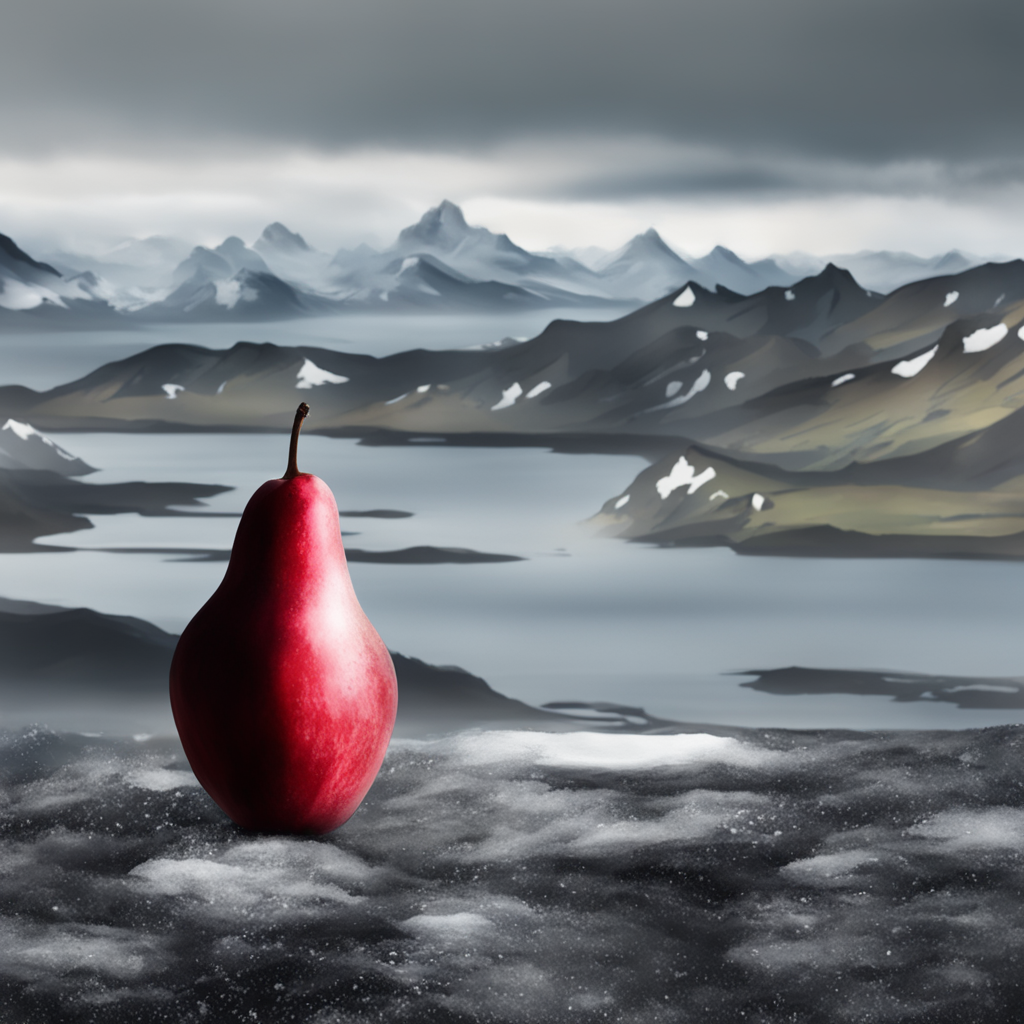}
    & \includegraphics[width=\linewidth]{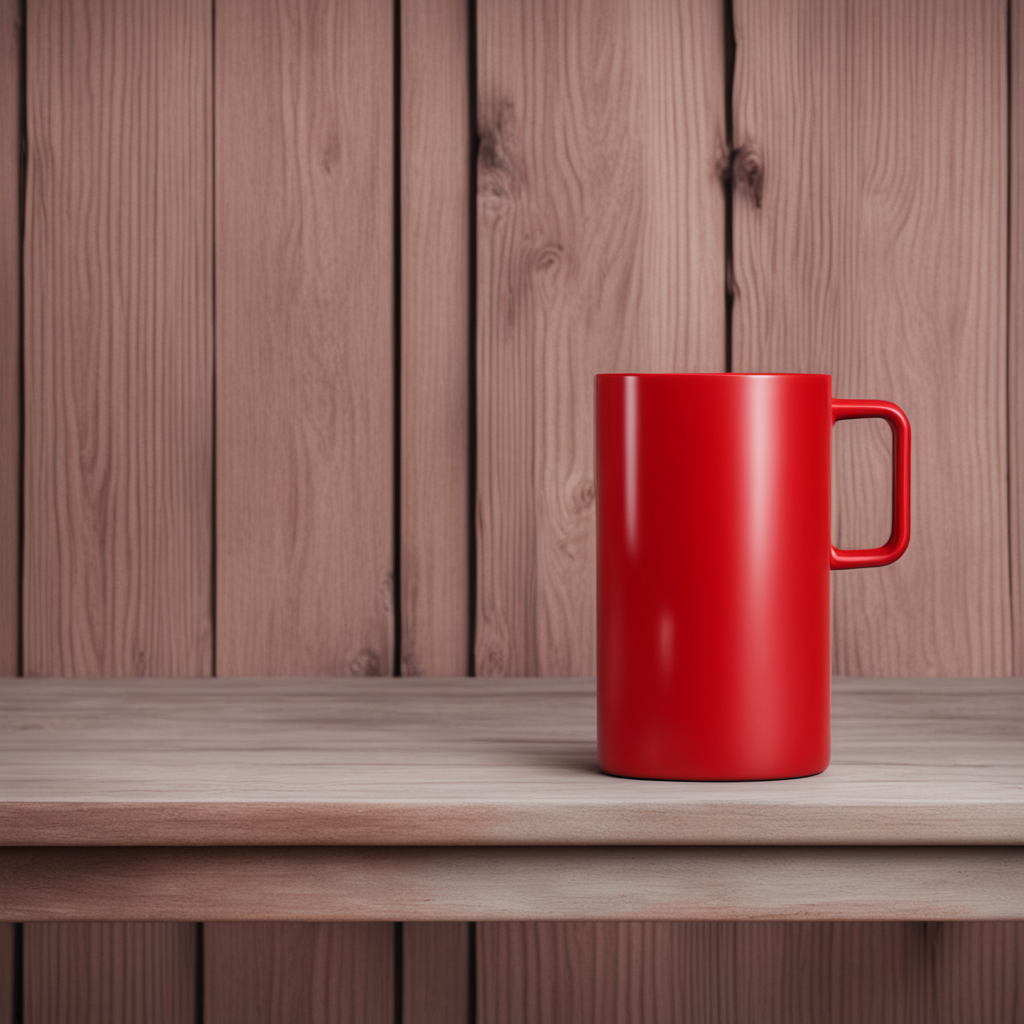}
    & \includegraphics[width=\linewidth]{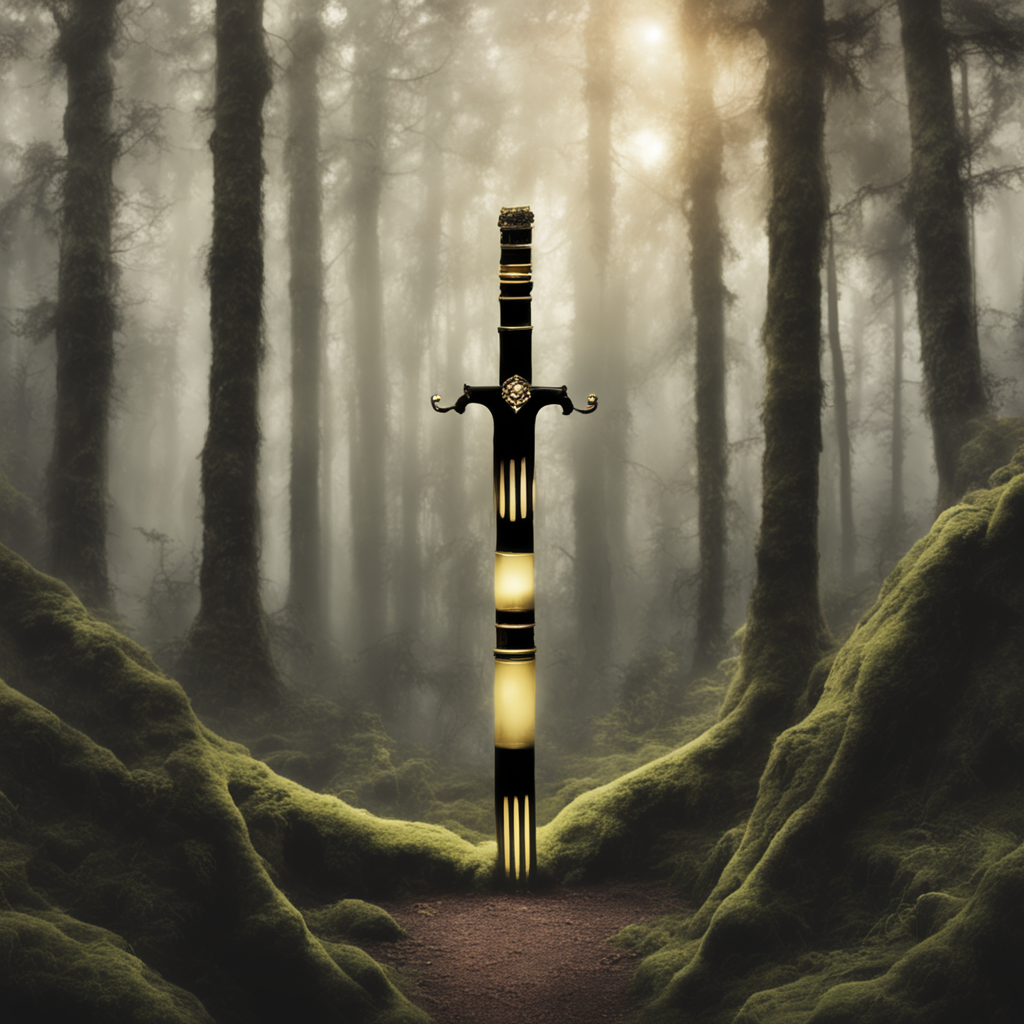}
    & \includegraphics[width=\linewidth]{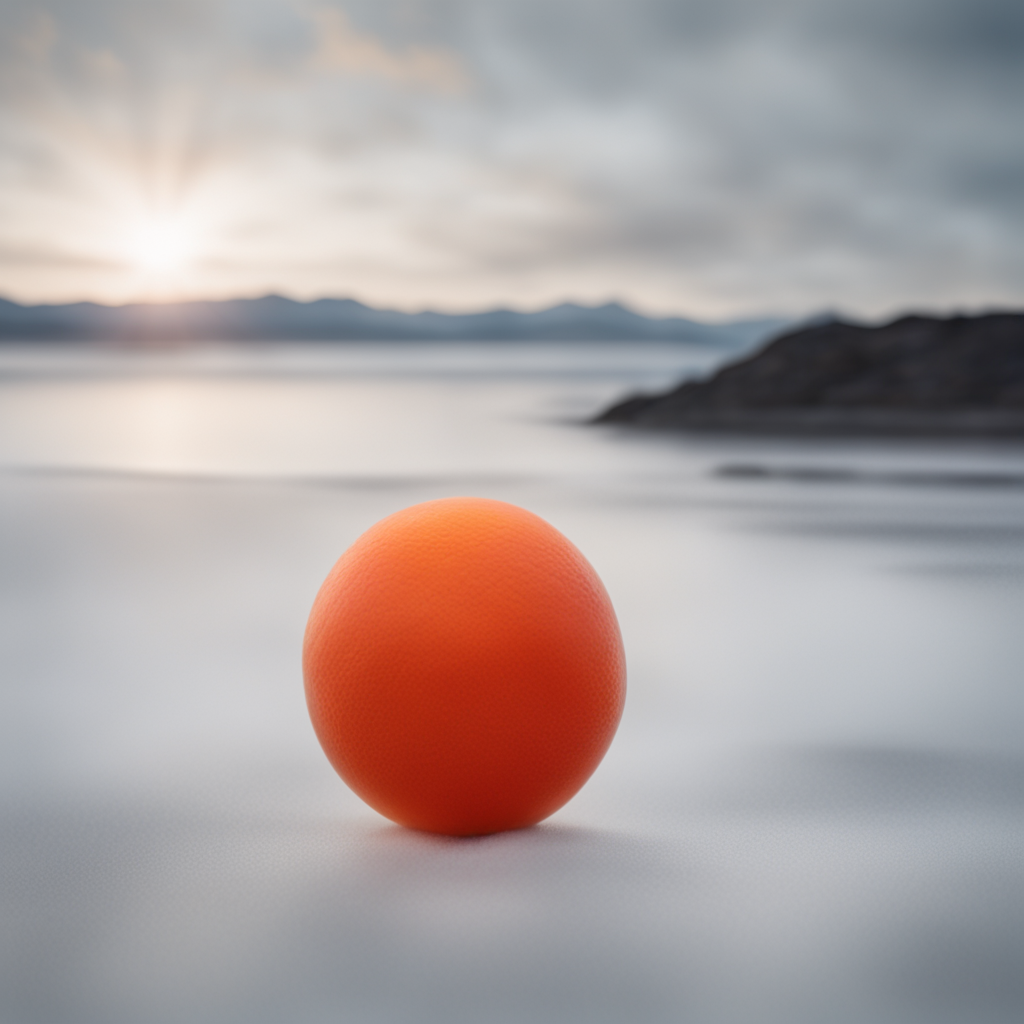}
    & \includegraphics[width=\linewidth]{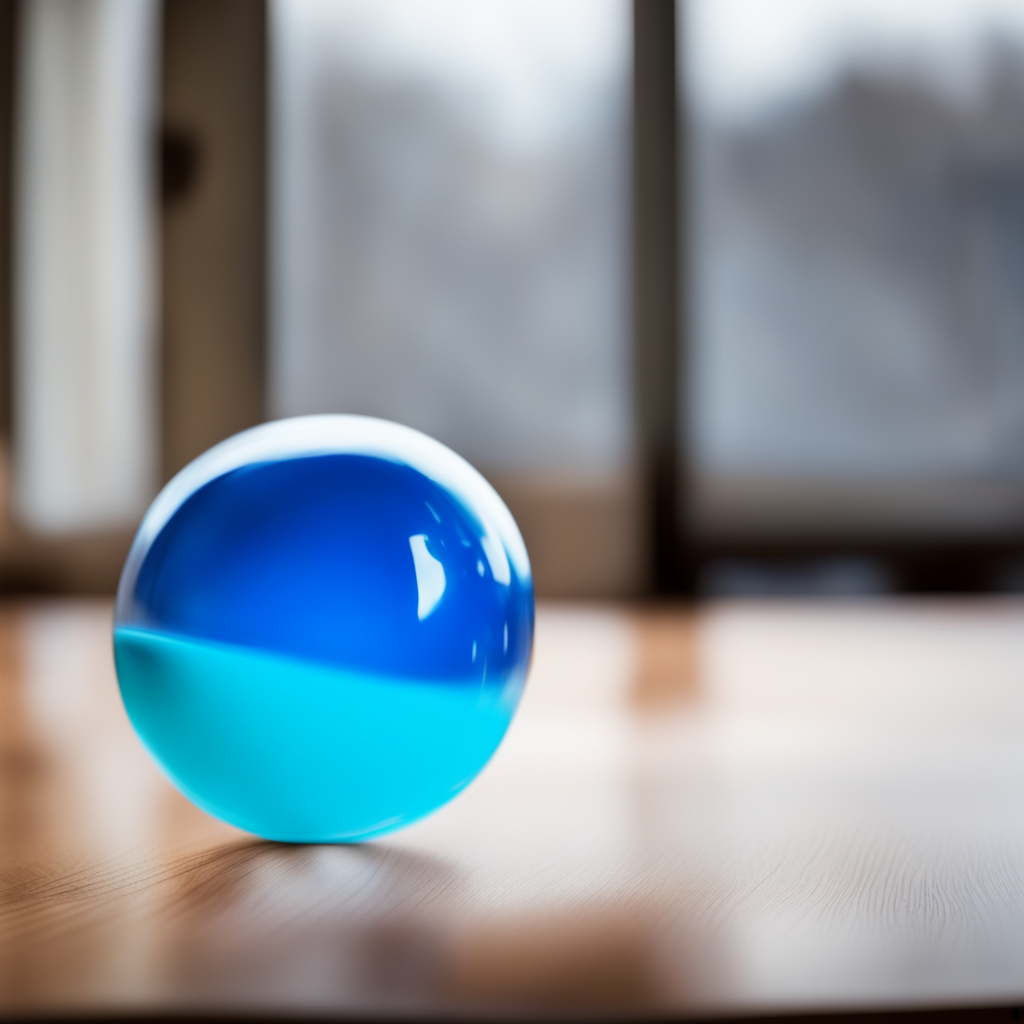}\\%[3pt]
\end{tabular}

%% MARCO
\caption{\textbf{Qualitative results.} We compare POEM with state-of-the-art image editing models across a diverse set of edit instructions, including geometric transformations (e.g., translation, scaling), appearance changes, and combinations of both. The specific prompts used are \textit{``Scale the bus by 0.56"}, \textit{``Move the pear left by 150px and make it red"}, \textit{``Scale the mug only vertically to 200px"}, \textit{``Make the sword gold"}, \textit{``Scale the orange by 2 and move it left by 150px"}, and \textit{``Move the ball left by 90px and make it blue"}.}
    % \caption{\textbf{Qualitative results.} We compare POEM to state-of-the-art image editing models. We test our edit instructions using translation, scaling, appearance changing, and a combination of them to showcase the precision of our pipeline. The prompts are from "Scale the bus to 0.56", "Move the pear left by 150px and make the pear red" , scale the mug only vertically by 200px", make the sword gold, scal ethe orange by 2.0, and move the orange left by 150px" move the ball to the left by 90px and make the ball blue/"}
    \label{fig:3-qualitative-comparison}
    \vspace{-1mm}
\end{figure}

\subsection{Visual Grounding}
\label{sec:4.2}

\mypar{Evaluation protocol.} 
% marco 
To assess the quality of the detected bounding box, we compute Intersection over Union (IoU) with the ground truth. If the MLLM fails to detect a bounding box, we fallback to a prediction covering the entire image. For images with multiple objects, we evaluate only the bounding box corresponding to the target object for transformation.

% original
% To assess the quality of the detected bounding boxes in the input image, we compute its Intersection over Union (IoU) with the ground truth bounding box. In the rare case that the MLLM fails to produce any bounding box, we use a fallback prediction that includes the entire image. When dealing with images containing multiple objects, we evaluate here only the bounding box that corresponds to the target object designated for transformation.

\mypar{Comparison.} 
We compare two MLLMs—Qwen2.5-VL~\cite{qwenQwen25TechnicalReport2025} and Intern-VL-2.5~\cite{chen2024internvl} in their 7B/8B and 72B variants. Model selection is guided by OpenCompass Open VLM leaderboard performance and dual H100 GPU compatibility.

% figure 4 - breaking point illustration 

\begin{figure}[t]
    \centering
\begin{tabular}{@{} 
  >{\centering\arraybackslash}m{0.166666667\linewidth} @{} 
  >{\centering\arraybackslash}m{0.166666667\linewidth} @{} 
  >{\centering\arraybackslash}m{0.166666667\linewidth} @{} 
  >{\centering\arraybackslash}m{0.166666667\linewidth} @{} 
  >{\centering\arraybackslash}m{0.166666667\linewidth} @{}
  >{\centering\arraybackslash}m{0.166666667\linewidth} @{}
  % >{\centering\arraybackslash}m{0.1\linewidth} @{}
  % >{\centering\arraybackslash}m{0.1\linewidth} @{}
  % >{\centering\arraybackslash}m{0.1\linewidth} @{}
  % >{\centering\arraybackslash}m{0.1\linewidth} @{}
  }
    % Header row (centered)
    \scriptsize Input image
    % & \scriptsize Resize to 80\%
    & \scriptsize Resize to 70\%
    % & \scriptsize Resize to 60\%
    & \scriptsize Resize to 50\%
    % & \scriptsize Resize to 40\%
    & \scriptsize Resize to 30\%
    & \scriptsize Resize to 10\%
    % & \scriptsize Resize to 20\%
    & \scriptsize Resize to 6.25\%\\%[3pt]
    
    % Example content rows
    \includegraphics[width=\linewidth]{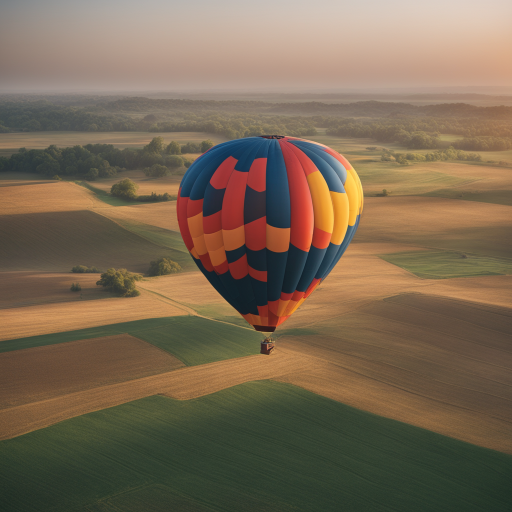}
    % & \includegraphics[width=\linewidth]{figures/BreakingPoint/evaluation_4_after_sld/sample_001.png}
    % & \includegraphics[width=\linewidth]{figures/BreakingPoint/evaluation_4_after_sld/sample_002.png}
    & \includegraphics[width=\linewidth]{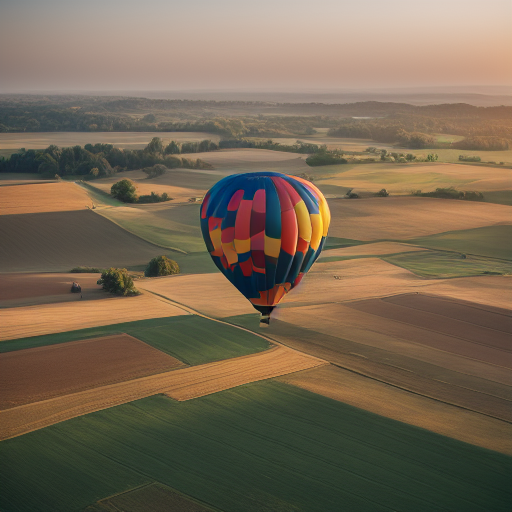}
    % & \includegraphics[width=\linewidth]{figures/BreakingPoint/evaluation_4_after_sld/sample_004.png}
    & \includegraphics[width=\linewidth]{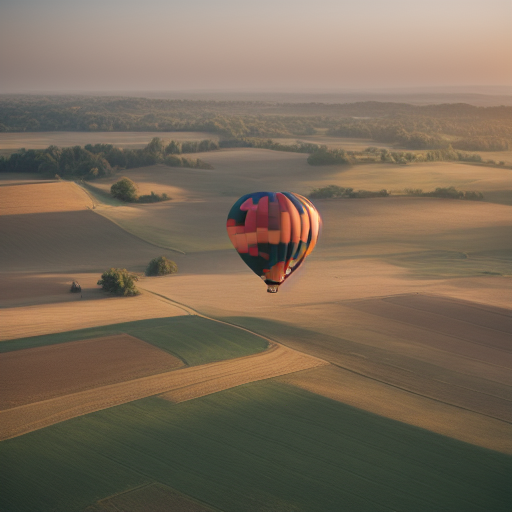}
    % & \includegraphics[width=\linewidth]{figures/BreakingPoint/evaluation_4_after_sld/sample_006.png}
    & \includegraphics[width=\linewidth]{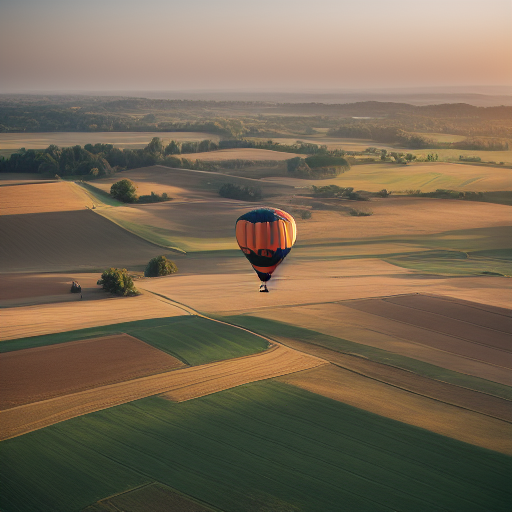}
    % & \includegraphics[width=\linewidth]{figures/BreakingPoint/evaluation_4_after_sld/sample_008.png}
    & \includegraphics[width=\linewidth]{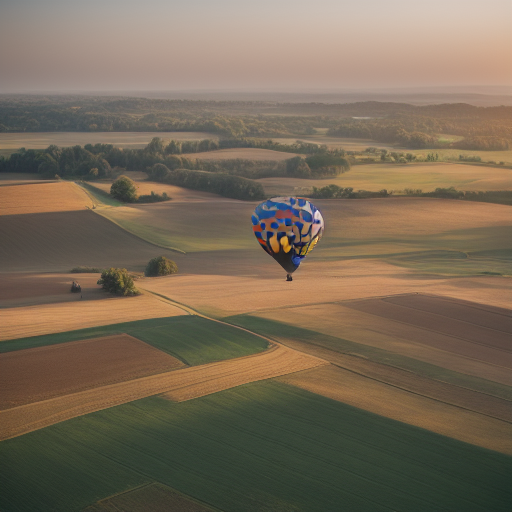} 
    & \includegraphics[width=\linewidth]{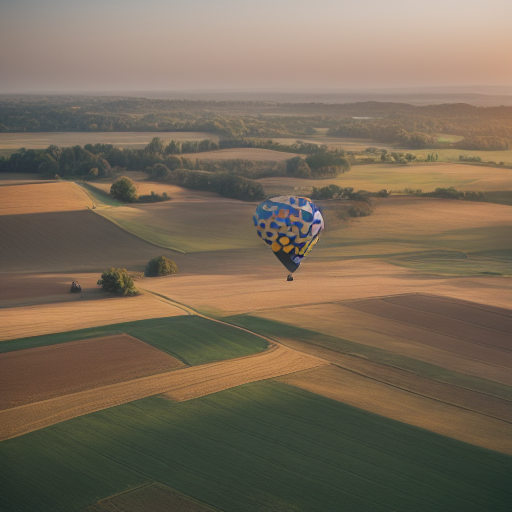} \\%[3pt]

        \scriptsize Input image
    & \scriptsize Move by 50px
    & \scriptsize Move by 100px
    & \scriptsize Move by 150px
    & \scriptsize Move by 200px
    & \scriptsize Move by 250px\\
    
    % Example content rows
    \includegraphics[width=\linewidth]{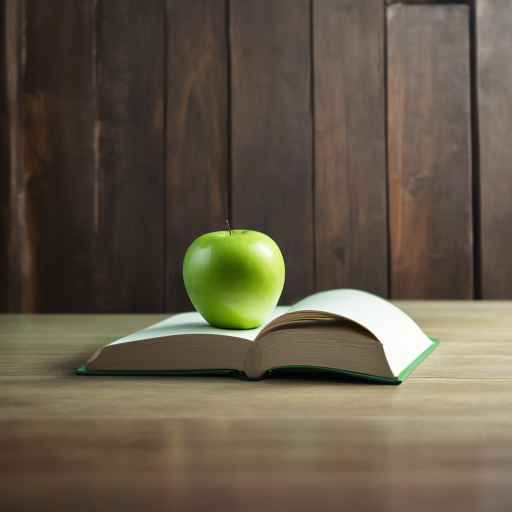}
    &     \includegraphics[width=\linewidth]{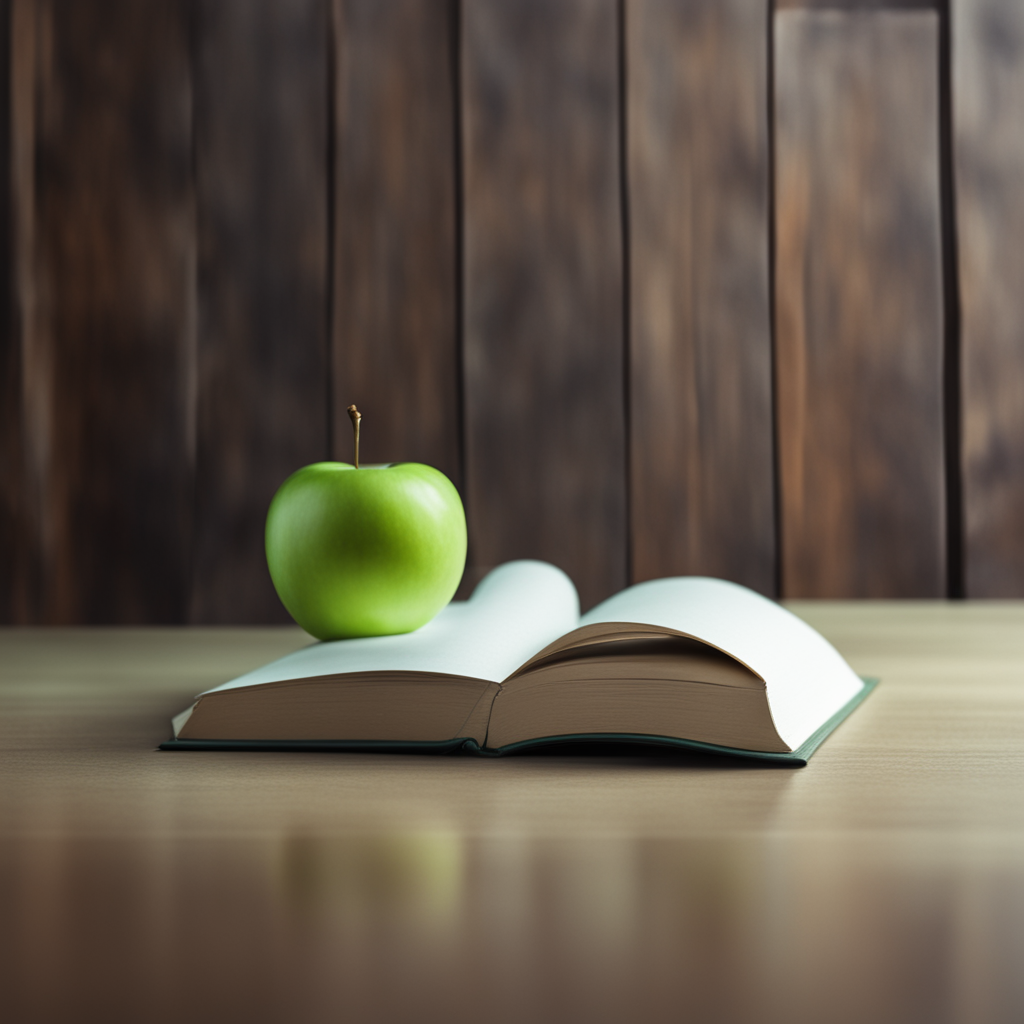}
    &     \includegraphics[width=\linewidth]{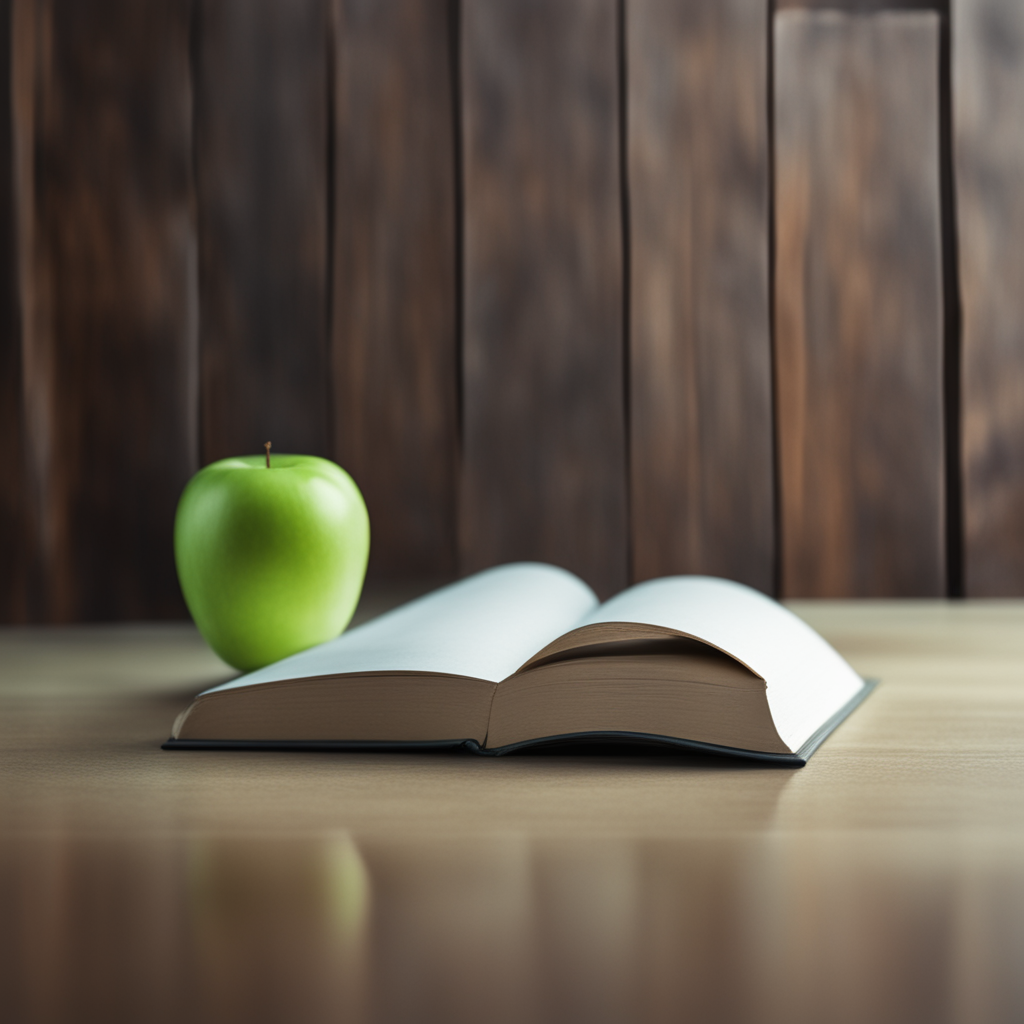}
    &     \includegraphics[width=\linewidth]{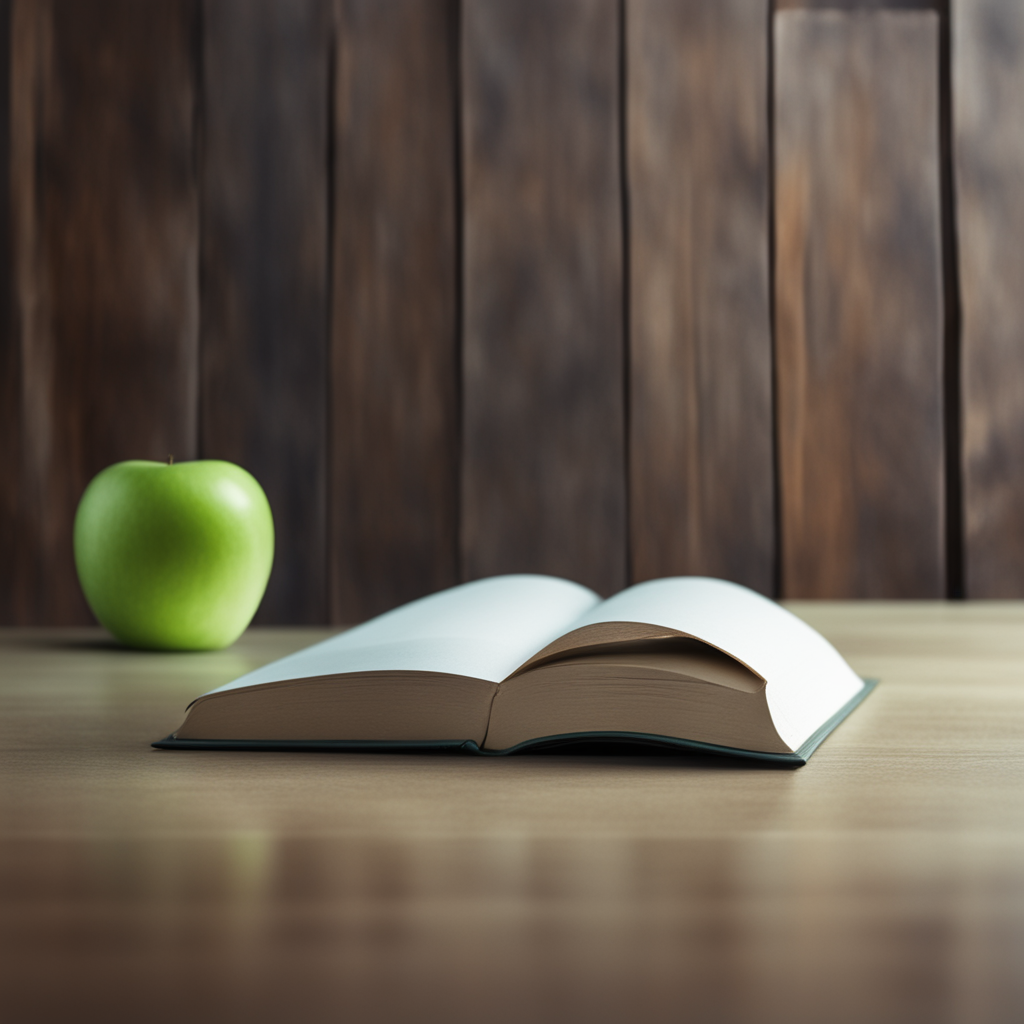}
    &     \includegraphics[width=\linewidth]{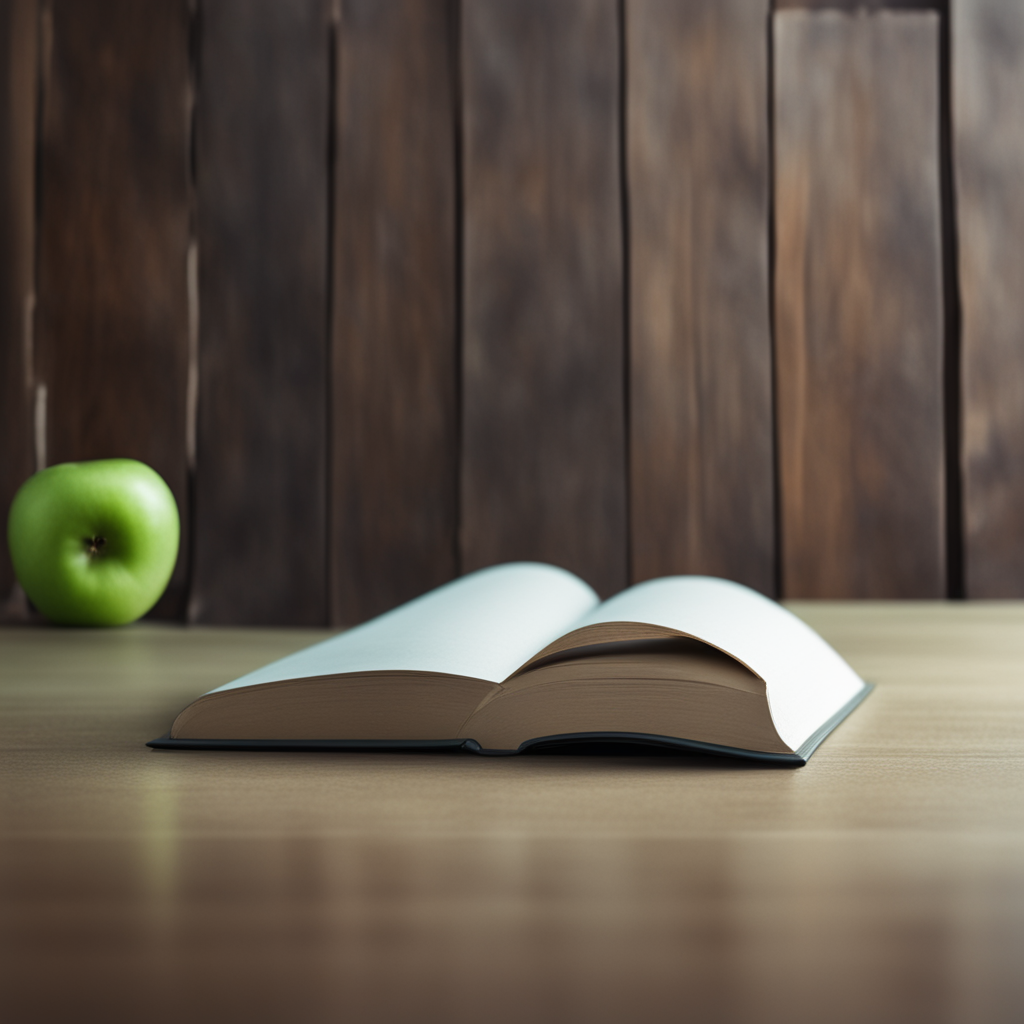}
    &     \includegraphics[width=\linewidth]{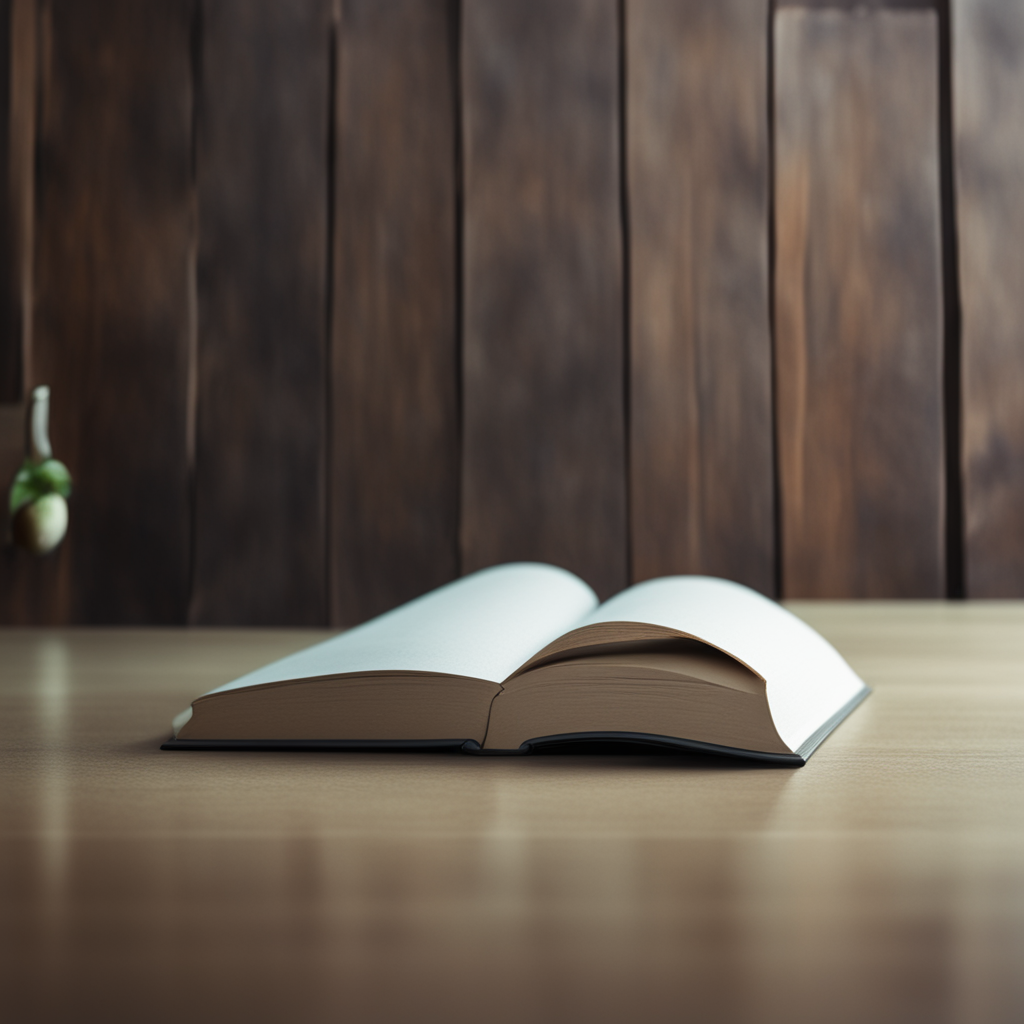} \\%[3pt]
\end{tabular}
 \caption{\textbf{Extreme transformations.} Although POEM's reasoning steps maintain robust mask quality and accurate transformation parameters, the image editing step~\cite{wuSelfcorrectingLLMcontrolledDiffusion2023} fails to generate an accurate image with an extreme edit (e.g., resizing fails at 10\%, and translation errors occur when the object approaches the image boundaries).
    % Despite robust mask quality and accurate computation of the transformation matrix-even under challenging conditions-the SLD pipeline (drawing step) fails to generate the accurate generated image when the prompt has a hard eedit, for example in the figure we notice that the resizing beings failing at 10\% and the translation 200px or more.
    }
    \label{fig:4-breaking-point}
\end{figure}

\mypar{Results.} QwenVL-7B  achieves an average IoU of 55.5\%, outperforming Intern-VL-8B by 38.1\% (Tab.~\ref{tab:1-main-quantitative}). This performance advantage is evident across all transformation categories. Considering their similar model sizes, these results highlight Qwen-VL’s superior effectiveness for this task. While InternVL-72B shows improved performance over its 8B variant, a similar trend is not observed for the QwenVL models. Therefore, we use QwenVL-7B for Visual Grounding in the remainder of our experiments.

\subsection{Detection Refinement}
\label{sec:4.3}

\mypar{Evaluation protocol.} We assess the segmentation quality by computing the IoU between the ground truth segmentation mask of the target object and the corresponding detected segmentation masks we obtain after the refinement stage. Similar to Sec.~\ref{sec:4.2}, when dealing with images containing multiple objects, we evaluate only the segmentation mask corresponding to the target object.

\mypar{Comparison.} We compare Grounded-SAM~\cite{renGroundedSAMAssembling2024} to SAM2~\cite{raviSAM2Segment2024}. Grounded-SAM is prompted with the predicted object class $c_i$ while SAM2 is prompted with the predicted segmentation point $s_i$. Both $c_i$ and $s_i$ are obtained from the MLLM.

\mypar{Results.} Grounded-SAM (denoted as G-SAM in Tab.~\ref{tab:1-main-quantitative}) exhibits a significant performance enhancement over SAM2 across all evaluated tasks, yielding an average IoU improvement of 56.9\%. These findings underscore the superior segmentation capabilities of Grounded-SAM over SAM2, particularly in refining object detection with greater accuracy and consistency.

\subsection{Edit Operation Parsing and Transformation}
\label{sec:4.4}

\mypar{Evaluation protocol.}
To assess transformation accuracy, we compute the ground-truth segmentation mask of the target object after applying the ground-truth transformation matrix.
We then measure the IoU between this mask and the predicted 
transformed mask $\hat{m}_{i*}$. This allows us to measure implicitly the error between our predicted transformation matrix $T$ and the ground-truth one. 

\mypar{Comparison.}
We evaluate two LLMs: Qwen2.5-Math-7B-Instruct~\cite{yangQwen25MathTechnicalReport2024}, which uses external tools like solvers and libraries, and DeepSeek-R1-Distill-Qwen-32B~\cite{deepseek-aiDeepSeekV3TechnicalReport2024}, relying on internal knowledge. Transformations are performed with OpenCV for geometric modifications. DeepSeek runs on a single NVIDIA H100 GPU (80GB), using up to 74GB of memory. We analyze two scenarios: (1) with our pipeline’s best models (see Sec~\ref{sec:4.3}) and (2) with an oracle ground-truth mask, isolating LLM-based reasoning effects. The second scenario measures transformation errors independently, while the first evaluates cumulative error from imperfect segmentation.
% We evaluate two large language models (LLMs): Qwen2.5-Math-7B-Instruct~\cite{yangQwen25MathTechnicalReport2024}, which uses external tools such as symbolic solvers and numerical libraries, and DeepSeek-R1-Distill-Qwen-32B~\cite{deepseek-aiDeepSeekV3TechnicalReport2024}, which relies on its internal knowledge. We perform transformations with OpenCV for precise geometric modifications. The DeepSeek model runs on a single NVIDIA H100 GPU (80GB), using a maximum of 74GB of memory. We analyze two scenarios about the segmentation mask of the input image: (1) using our pipeline's best grounding and refinement models (see Sec~\ref{sec:4.3}) and (2) using an oracle ground-truth mask, helping us isolate the impact of LLM-based reasoning. 
% The second scenario allows us to measure transformation errors independently, while the first assesses the cumulative error introduced by imperfect segmentation masks.

\mypar{Results.}
QwenMath (denoted as QwenM in Tab.~\ref{tab:1-main-quantitative}) consistently outperforms DeepSeek across all transformation categories by a clear margin, achieving 7-41\% higher IoU scores.
This is likely due to QwenMath's tool-integrated-reasoning approach, which enhances matrix multiplication accuracy for complex transformations. The same trend appears for both evaluation scenarios (i.e., predicted and oracle masks). 
Using oracle masks improves IoU scores from 49.2\% to 55.6\%, suggesting that the primary source of error comes from the transformation prediction rather than segmentation inaccuracies. This highlights the importance of robust mathematical reasoning in object-level transformations.

\subsection{Edit Guided Image-to-Image Translation}
\label{sec:4.5}

\mypar{Evaluation protocol.} 
To assess the image editing quality, we go beyond standard image quality metrics (e.g., FID), and instead, we measure the alignment of the edited images with the input prompts and the transformations. Specifically, we first use Grounded SAM to estimate the segmentation mask of the transformed object in the edited image. We then compute the IoU between this mask and the segmentation mask after applying the ground-truth transformation.

\mypar{Comparison.} 
We use the Stable Diffusion v2.1~\cite{rombachHighResolutionImageSynthesis2022} as our pre-trained diffusion model and adopt the latent space operations from~\cite{wuSelfcorrectingLLMcontrolledDiffusion2023}. Additionally, we experiment with Stable Diffusion XL~\cite{podellSDXLImprovingLatent2023} as a refiner to improve the image quality.

\mypar{Results.} Comparing the two strategies for generating the final image, we observe minimal performance differences, with SDXL refinement leading to an average IoU drop of only 0.8\%. This change is statistically insignificant, but qualitatively, the refined images exhibit improved visual quality. 
When comparing this IoU accuracy from this step with the one from the previous section, we observe a significant 10.8\% drop (from 49.2\% to 38.4\%). This drop is caused by the image editing process, which does not always fully adhere to the guided segmentation masks. In Sec.~\ref{sec:4.7} and Fig.~\ref{fig:4-breaking-point}, we further analyze these image editing limitations, particularly in cases with extreme transformations.

% As presented in Tab.~\ref{tab:1-main-quantitative}, we observe that the results remain largely consistent, with only a slight average deterioration of approximately 0.8 percentage points when utilizing the refiner. This change is statistically non-significant; however, qualitatively, the refined images exhibit improved visual quality. It is noteworthy that the mask performance shows a minor decline. Overall, the shear operation yields the best-case scenario across all results, while the reasoning task produces the least favorable results, as it is intrinsically more complex. 

% \subsection{Qualitative comparison with text-based image editing} 
% \label{sec:4.6}

%% Marco

\mypar{Comparison to state-of-the-art image editing.} 
Fig.~\ref{fig:3-qualitative-comparison} shows a qualitative comparison of POEM with state-of-the-art models, including IP2P~\cite{brooksInstructPix2PixLearningFollow2023}, LEDITS++~\cite{brackLEDITSLimitlessImage2024}, and TurboEdit~\cite{deutch2024turboedit}. The figure demonstrates POEM’s ability to generate more faithful, targeted edits. Tab.~\ref{tab:1-main-quantitative} reports quantitative comparisons, where POEM achieves 38.4\%, surpassing IP2P (34.4\%), TurboEdit (33.8\%), and LEDITS++ (35.0\%) by about 3\%. POEM excels in \textit{translate} and \textit{scale} operations, with improvements from 27.4\% to 32.6\% and 32.9\% to 39.7\%, respectively. These results highlight our model’s superior performance, producing more precise edits and accurate transformation parameters that better align with the user's intended modifications.

\subsection{Limitations}
\label{sec:4.7}

While POEM achieves precise object-level transformations, the image editing step inherited from diffusion models has certain limitations.

First, when dealing with extreme transformation, POEM can predict accurate parameters, but diffusion models struggle to generate objects that become too small relative to the image size. This issue is most pronounced when objects shrink to less than 10\% of their original size or move partially outside the image boundaries (Fig.~\ref{fig:4-breaking-point}).
To measure this effect quantitatively, we categorize the transformations of our dataset into easy, medium, and hard based on the IoU difference between the original and transformed masks. After applying the LLM-based step, we obtain an IoU of 68\% for easy, 66\% for medium and 40\% for hard transformations. In contrast, the image editing step lowers IoU to 55\%, 54\%, and only 30\%, respectively, highlighting challenges in handling severe modifications.

Second, our approach currently focuses on rigid-body transformations, as our editing step~\cite{wuSelfcorrectingLLMcontrolledDiffusion2023} does not support non-rigid deformations, such as altering human poses (e.g., raising an arm). 
A possible solution is integrating more explicit control signals, similar to Self-Guidance~\cite{epsteinDiffusionSelfGuidanceControllable2023}. 
However, Self-Guidance is very sensitive to hyperparameters which impacts its reproducibility and generalization ability. Future work could refine its framework to ensure reliable image edits, such as preserving background integrity across transformations~\cite{epsteinDiffusionSelfGuidanceControllable2023}.

%% file: 5_conclusion.tex
\section{Conclusion}

We proposed POEM, a novel approach that leverages MLLMs, LLMs, and segmentation models to enhance image editing capabilities through precise text-instruction-based operations. Our approach facilitates object-level editing by generating accurate masks alongside relevant contextual information derived from the input image. This feature empowers users to perform precise modifications directly from natural language instructions.
Additionally, we introduced VOCEdits, a comprehensive dataset designed for evaluating object-level editing, which establishes a robust benchmark for tasks related to detection, transformation, and synthesis. 
By integrating MLLMs with diffusion models, POEM bridges the gap between high-level instructional reasoning and low-level spatial control, laying the foundation for future research in multimodal image editing. 
We believe our work will drive advancements in controllable image synthesis, making precise and intuitive editing more accessible to users. 

\noindent\textbf{Acknowledgments.}
D. Papadopoulos was supported by the DFF Sapere Aude Starting Grant ``ACHILLES". This work was partly supported by the Pioneer Centre for AI, DNRF grant number P1.
We would like to thank Thanos Delatolas for the insightful discussions.

% Onur: no future direction in the conclusion
% One promising direction for future work is the integration of precise text-based object-level image editing into an iterative image generation and refinement pipeline, leveraging the strengths of MLLMs. Current image generation models often struggle with complex prompts, such as ``create a chessboard where white is checkmated," despite their apparent simplicity. MLLMs, however, excel at interpreting such nuanced instructions and can identify errors in generated images, proposing precise textual edits for refinement. By developing image editing models capable of understanding and executing these precise text-based prompts, we can create a seamless feedback loop between generation and refinement. While approaches, such as SLD \cite{wuSelfcorrectingLLMcontrolledDiffusion2023}, aim to establish more complicated interfaces for communication between LLMs and image editing tools, text-based precise prompts offer a more natural and intuitive solution for MLLM-to-image-model interaction. This advancement would bridge the gap between high-level semantic understanding and precise pixel-level control, significantly enhancing the quality and accuracy of image generation and editing.